\documentclass[10pt,journal,compsoc]{IEEEtran}
\usepackage[latin1]{inputenc}
\usepackage{amsmath}
\usepackage{amsfonts}
\usepackage{epsfig}
\usepackage{graphicx}
\usepackage{physics}
\usepackage{algorithm}
\usepackage{algorithmic}
\usepackage{bm}
\usepackage{amssymb}
\usepackage{booktabs,colortbl,multirow}
\usepackage{subfig}
\usepackage[numbers,sort&compress]{natbib}
\usepackage{afterpage}
\usepackage{dblfloatfix}

\def\etal{\emph{et al.~}}

\begin{document}

	
	\title{Measuring the Temporal Behavior of Real-World Person Re-Identification}
	\author{Meng Zheng, \IEEEmembership{Student Member, IEEE,}
		Srikrishna Karanam, \IEEEmembership{Member, IEEE,} \\
		and Richard J.~Radke, \IEEEmembership{Senior Member, IEEE}
		\IEEEcompsocitemizethanks{\IEEEcompsocthanksitem M.~Zheng and R.J.~Radke are with the Department
			of Electrical, Computer, and Systems Engineering, Rensselaer Polytechnic Institute, Troy, NY 12180 USA
			(e-mail: zhengm3@rpi.edu, rjradke@ecse.rpi.edu).}
		\IEEEcompsocitemizethanks{\IEEEcompsocthanksitem S.~Karanam is with Siemens Corporate Technology, Princeton, NJ 08540 USA
			(e-mail: srikrishna.karanam@siemens.com).}
		\IEEEcompsocitemizethanks{\IEEEcompsocthanksitem Corresponding author: M.~Zheng.  This material is based upon work supported by the U.S.~Department of Homeland Security under Award Number 2013-ST-061-ED0001. The views and conclusions contained in this document are those of the authors and should not be interpreted as necessarily representing the official policies, either expressed or implied, of the U.S.~Department of Homeland Security.}}

\IEEEtitleabstractindextext{
\begin{abstract}
		Designing real-world person re-identification (re-id) systems requires attention to operational aspects not typically considered in academic research. Typically, the probe image or image sequence is matched to a gallery set with a fixed candidate list. On the other hand, in real-world applications of re-id,  we would search for a person of interest in a gallery set that is continuously populated by new candidates over time. A key question of interest for the operator of such a system is: how long is a correct match to a probe likely to remain in a rank-k shortlist of candidates? In this paper, we propose to distill this information into what we call a Rank Persistence Curve (RPC), which unlike a conventional cumulative match characteristic (CMC) curve helps directly compare the \textit{temporal} performance of different re-id algorithms. To carefully illustrate the concept, we collected a new multi-shot person re-id dataset called RPIfield. The RPIfield dataset is constructed using a network of 12 cameras with 112 explicitly time-stamped actor paths among about 4000 distractors. We then evaluate the temporal performance of different re-id algorithms using the proposed RPCs using single and pairwise camera videos from RPIfield, and discuss considerations for future research.
	\end{abstract}
    }
\maketitle
	
	\section{Introduction}
	\label{sec:intro}
	Research in the area of automatic person re-identification, or re-id, has exploded in the past ten years. The re-id problem is generally stated as: given images of a person of interest as seen in a ``probe" camera view, how can we find the same person among a set of candidate people seen in a ``gallery" camera view? Re-id research to date typically falls into one or more of the following categories:
	\begin{itemize}
		\item Appearance modeling, in which the goal is to design or learn feature representations invariant to challenges like viewpoint and illumination variation (e.g., \cite{Loris_CVIU13, RCNN_CVPR16, descrip_CVPR13,li2017learning,zhao2017spindle}).
		\item Metric learning, in which the goal is to learn, in a supervised fashion, a distance metric that is used to search for the person of interest in the gallery set (e.g., \cite{KISSME_CVPR12, LOMO_XQDA_CVPR15, LFDA_CVPR13, NFST_CVPR16, chen2017beyond}).
		\item Multi-shot re-id, in which both the probe and gallery candidates are represented as short image sequences/video clips instead of single frames (e.g., \cite{BlockSparse_17,multishot_yang_BMVC15,WHOS_PAMI15,you2016top,zhou2017see}).
	\end{itemize}

\begin{figure*}[h!]
		\centering
		\includegraphics[width=.95\linewidth]{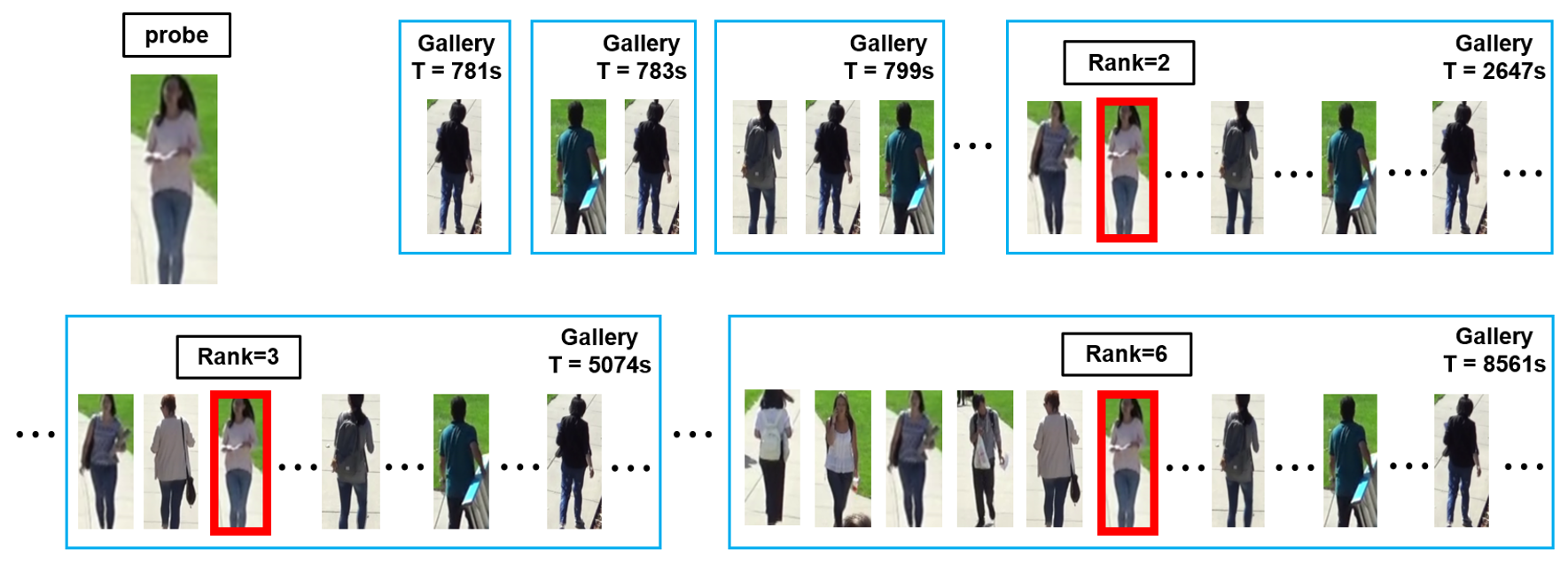}%
		\caption{An example showing the temporal evolution of a gallery set captured in our RPIfield dataset, and the corresponding time-varying rank for a given query achieved by a re-id algorithm. The candidate arrives at t = 2647s at rank 2, and drops in rank over time as more candidates arrive in the target camera.} 
		\label{fig:intro_demo}
	\end{figure*}
    
	We refer the reader to recent experimental and algorithmic surveys by Karanam \etal \cite{benchmark_17} and Zheng \etal \cite{zheng2016person}. While these issues are all critical for designing successful re-identification algorithms, research in these areas generally oversimplifies the problem that would face a real-world user of a deployed re-id system. In particular, the temporal aspect of the re-id problem is ignored in most academic re-id research. That is, in the real world, candidates would be constantly added to the gallery as new subjects are automatically tracked, as opposed to presented to an algorithm all at once. Even if a correct match to the probe appears in a rank-ordered shortlist shortly after they appear in a gallery camera, this isn't helpful to a user if the candidate is immediately pushed off the list after a few minutes by a new wave of incoming candidates. Figure \ref{fig:intro_demo} illustrates an example in which the person of interest first appears at $T=2647s$ at rank 2, but the rank drops to 6 at $T = 8561s$ as more candidates are added to the gallery over time. To the user, a natural question is, how long can a correct match be expected to stay in the shortlist under typical circumstances? To this end, we take a broader view of how re-id algorithms perform, judging them on this notion of persistence in time and not just raw batch performance as typically presented in a Cumulative Match Characteristic (CMC) curve. This is particularly appealing in areas where re-id finds the most immediate and practical application, such as crime detection or prevention. The perpetrator of a crime may re-appear after a significantly long time interval (e.g., several days or even months), and it is important to understand how existing re-id algorithms perform at this time scale. 
	
	In this paper, we study several temporal aspects of re-id and propose new evaluation methodologies that allow different re-id algorithms to be compared based on a concept we call \emph{rank persistence}. We discuss strategies for evaluating algorithms in circumstances when the same person of interest can appear multiple times in the gallery as well as when the performance on multiple persons of interest should be aggregated. The key concept we propose is called the Rank Persistence Curve (RPC), which quantifies the percentage of candidates that remain at a certain rank for a given duration. RPCs for different algorithms can be directly compared on the same camera gallery to allow a user to make informed choices about expected performance in real-world deployments.	
	
	
	To help evaluate the temporal performance of re-id algorithms under the considerations discussed above, we need datasets that come with explicitly time-stamped images. Current re-id benchmarking datasets such as VIPeR \cite{VIPeR_ECCV08}, iLIDSVID \cite{iLIDS-VID_ECCV14}, and MARS \cite{MARS_ECCV16} lack the kind of time stamps for the gallery sets needed to understand the full potential of the RPC concept, rendering them inappropriate for our use here. While it might be possible to repurpose them for temporal re-id research, adding artificial time stamps seems suboptimal. To this end, we propose a new multi-shot multi-camera dataset, named RPIfield, that includes multiple reappearances of 112 ``actors" walking along specified paths, among almost 4000 distractor pedestrians. We illustrate RPCs with examples and experiments drawn from our dataset and then compare the temporal performance of several re-id algorithms. We conclude the paper with further discussion of the temporal implications of re-id, and suggestions for future research in this area. The RPIfield dataset and RPC code will be made publicly available upon acceptance of this paper.  A preliminary version of this work appeared in Karanam \etal \cite{Srik_ICDSC17}. 
	
\section{Related Work}
Re-id is a widely studied field; we refer the reader to the recent experimental survey by Karanam \etal \cite{benchmark_17} and the algorithmic survey by Zheng \etal \cite{zheng2016person} for an extensive discussion. Here we only discuss work directly related to the problems we address in this paper- temporally rich datasets and performance evaluation. Existing re-id algorithms are typically evaluated on academic re-id datasets such as VIPeR \cite{VIPeR_ECCV08}, Market1501 \cite{Market1501_ICCV15}, CAVIAR \cite{CAVIAR_BMVC11}, PRID \cite{PRID_SCIA11}, 3DPeS \cite{3DPeS_ACM11}, WARD \cite{WARD_ECCV08}, iLIDSVID \cite{iLIDS-VID_ECCV14} and RAiD \cite{RAiD_ECCV14}. These datasets are specifically hand-curated to only have sets of bounding boxes for the probes and the corresponding matching candidates in the gallery. On the other hand, re-id is typically only a small module in a larger end-to-end surveillance system that tracks people in camera networks \cite{Octavia_CSVT16}. Consequently, it is important to construct datasets that help evaluate this critical module from a real-world operational perspective, and as described in the previous section, the temporal aspect is at the forefront. In the past, there have been some efforts to this end. 

Figueira \etal proposed the HDA+ \cite{figueira2014hda} dataset as a testbed for evaluating an automatic re-id system. Fully labeled frames for 30-minute long videos from 13 disjoint cameras were provided as part of this dataset. Ristani \etal presented a much larger-scale multi-camera dataset, called DukeMTMC4ReID \cite{DukeMTMC_CVPR17}, where images corresponding to 1852 unique people were captured from a disjoint 8-camera network. While these, and other relevant datasets described further in Karanam \etal \cite{benchmark_17}, are multi-shot and multi-camera by design, they lack the crucial temporal aspect we describe in this work. This is a key difference between the proposed RPIfield dataset and others. Specifically, RPIfield comes with explicit time-stamp information about when a particular person re-appeared. Most people captured in our dataset have multiple re-appearances, and each re-appearance is time-stamped. 
	
	
	A very popular evaluation metric in the field of re-id is the cumulative match characteristic (CMC) curve, which is a plot of the re-id rate at rank-k \cite{benchmark_17}. However, the CMC curve completely disregards the time at which a particular person appeared in the gallery set, instead using the entire gallery set as a batch to compute the re-id performance. Some recent work also uses the mean average precision (mAP) \cite{Market1501_ICCV15} to evaluate algorithms on multi-shot datasets with multiple ground-truth correct matches of the person of interest in the gallery. However, mAP has the same fundamental problem as the CMC; the assumption is that the entire gallery set is available at the same time, which is not how a gallery set would be structured in the real world. To bridge this gap between academic re-id evaluation and usability for real-world systems, we present the concept of rank persistence, where we explicitly ask the question: how long will a person of interest stay in the top-k retrieved list as more and more candidates are populated in the gallery over time? 
    

	

	
	\section{The RPIfield Dataset}
	\label{sec:dataset}
	As discussed in Section \ref{sec:intro}, time-stamped datasets are critical for evaluating real-world re-id systems that detect and track people automatically. Here, in order to simulate such real-world requirements, we present a new multi-shot multi-camera re-id dataset, called RPIfield.
	
\begin{figure}[h!]
		\centering
		\includegraphics[width=0.9\linewidth]{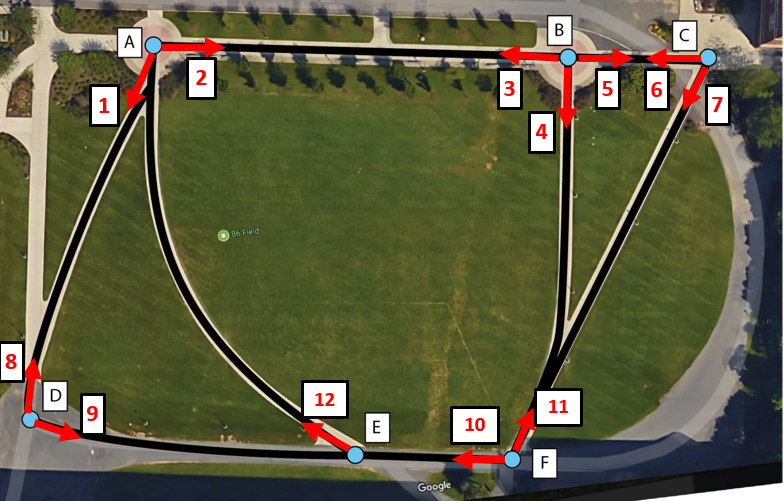}%
		\caption{Overhead view of the RPIfield camera locations and orientations, superimposed on a map of the RPI '86 Field.}
		\label{fig:camera}
	\end{figure}

	RPIfield is constructed from 12 synchronized cameras placed around an outdoor field on the campus of Rensselaer Polytechnic Institute. Figure \ref{fig:camera} shows the camera network layout. 6 poles were positioned around the field, one each at points A through F. Each red arrow represents a camera with its corresponding viewing direction.
	
	\begin{figure*}[!h]
		\centering
		\subfloat[Cam ID: 3, T=668.1s]{{\includegraphics[width=.3\linewidth,height=3cm]{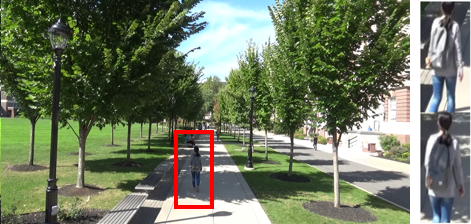} }} \quad
		\subfloat[Cam ID: 2, T=738.1s]{{\includegraphics[width=.3\linewidth,height=3cm]{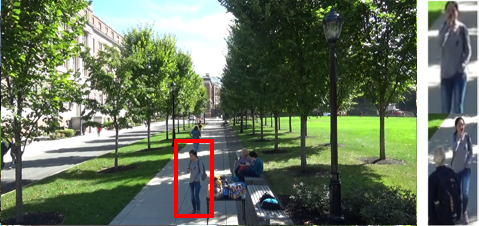} }}\quad
		\subfloat[Cam ID: 1, T=767.7s]{{\includegraphics[width=.3\linewidth,height=3cm]{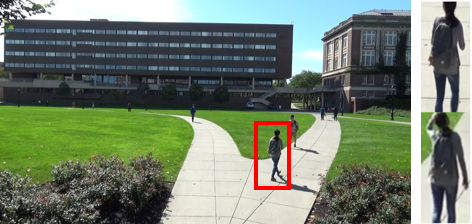} }}
		\\
		\subfloat[Cam ID: 4, T=1133.1s]{{\includegraphics[width=.3\linewidth,height=3cm]{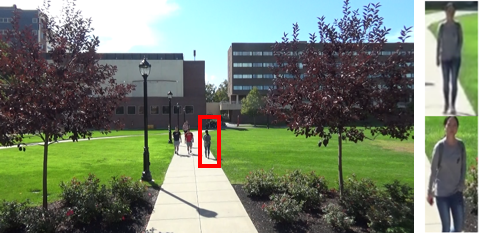} }}\quad
		\subfloat[Cam ID: 5, T=1330.9s]{{\includegraphics[width=.3\linewidth,height=3cm]{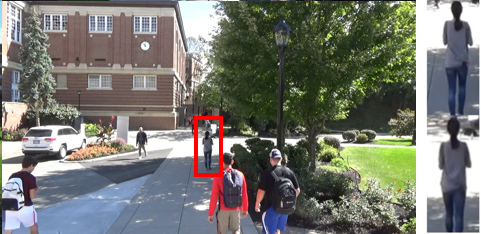} }}\quad
		\subfloat[Cam ID: 6, T=1343.8s]{{\includegraphics[width=.3\linewidth,height=3cm]{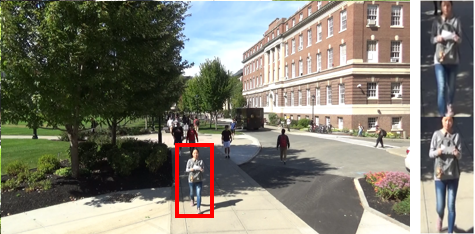} }}%
		\\
		\subfloat[Cam ID: 7, T=1365.7s]{{\includegraphics[width=.3\linewidth,height=3cm]{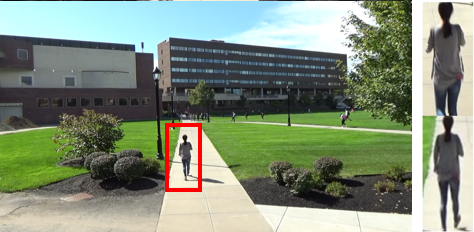} }}\quad
		\subfloat[Cam ID: 10, T=1441.0s]{{\includegraphics[width=.3\linewidth,height=3cm]{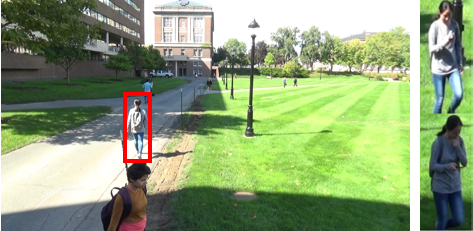} }}\quad
		\subfloat[Cam ID: 8, T=1523.7s]{{\includegraphics[width=.3\linewidth,height=3cm]{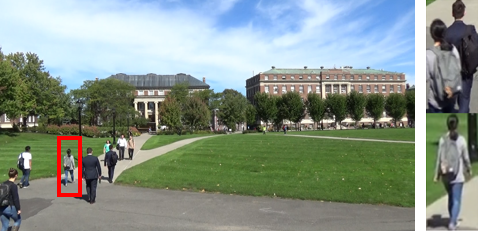} }}%
		\caption{Illustration of the reappearances of Participant 1 in different camera views. T is the time for each reappearance. Figures (a) to (i) are placed in time order of the appearance of Participant 1. The right column of each subfigure shows sample automatically extracted bounding boxes used as input to the re-id algorithms.} 
		\label{fig:demo_dataset}
	\end{figure*}
	
	
	The total length of videos from the collected camera feeds is about 30 hours, with each individual video being about 150 minutes in duration. The videos were recorded at full high-definition (1080p) resolution, with each camera's frame rate being 30fps. We collected all data between 11:30 AM to 2 PM on a weekday to capture the typically heavy traffic associated with this particular time period. This was also important to capture a large number of distractors to simulate a realistic re-id scenario. Cameras that have opposite viewing directions along the same path (e.g., camera pairs 1 and 8, or 5 and 6) have some overlap in their camera views. All other cases involve non-overlapping views. The images captured comprise an eclectic mix of challenges including viewpoint variations, intra- and inter-camera illumination variations, and occlusions.
	
	In our dataset, there are 112 known participants (actors) in total, with each showing up in at least 3 camera views. The participants were asked to walk along specific, pre-defined paths (provided by us) between different points (A through F) around the field shown in Figure \ref{fig:camera}. In order to ensure re-appearances of the actors in all camera views, each assigned path for a participant contains at least 3 different points. For example, if a participant is walking along the path $A \rightarrow B\rightarrow C\rightarrow F\rightarrow B \rightarrow A$, s/he will appear in camera views 2, 3, 5, 6, 7, 11, 4, 3 and 2. Each participant was assigned a different walking path. During this process, we also captured images of all other pedestrians (distractors) walking along all the paths, which form the candidate set in each of the 12 galleries. An illustrative example of multiple re-appearances of a certain participant in multiple camera views is shown in Figure \ref{fig:demo_dataset}. Each subfigure in Figure \ref{fig:demo_dataset} shows the appearance of the participant in a different camera at a different time. 
	
	To automatically collect image sequences for each person, we used an off-the-shelf person detector, based on the aggregated channel features (ACF) algorithm of Dollar et al.~\cite{ACF_BMVA10,ACF_PAMI14} to crop individual images in each frame. After collecting all the bounding boxes of all the detected people (including participants and distractors), we first use the intersection over union (IoU) of the bounding boxes to get a collection of raw tracklets. Specifically, for each cropped bounding box, we calculate the IoU of the current bounding box with all bounding boxes collected in the previous 90 frames, with a threshold of 0.4. We then match the current bounding box with the identity of the frame image achieving the largest IoU value. For bounding boxes with no surviving IoUs, we assign a new label. Since there will be false detections and broken tracklets (e.g., multiple tracklets for the same person) due to illumination variations and occlusions, we manually corrected all these errors to ensure each unique image sequence corresponds to one appearance of a person. The time stamp information for each image is preserved through the corresponding frame number. Sample images of Participant 1 in multiple camera at different times are shown to the right of each subfigure in Figure \ref{fig:demo_dataset}.
	
	A statistical summary of RPIfield is shown in Table \ref{table:stat}, where \#BBx. is the total number of person images generated by the manual annotation process after applying the ACF person detector, \#Par. is the number of known actors in each camera view, \#Reapp. is the total number of re-appearances (excluding first appearance) of the known actors in each camera view, \#Ped. is the number of distractors in each camera view, and \#Seq. is the number of collected image sequences. \#Len. is the length of each camera videos given in minutes. The last row provides an aggregated picture for the camera network used to construct RPIfield. 

	\begin{table}[!h]
		\caption{Statistics of the RPIfield dataset.}
		\centering
		\scalebox{1.0}{
			\setlength{\extrarowheight}{.2em}
			\begin{tabular}{|c|c|c|c|c|c|c|}
				\hline
				Cam No. & \#BBx. & \#Par. & \#Reapp. & \#Ped. & \#Seq. & Len.\\
				\hline
				Cam 1 & 59,230 & 78 & 107 & 297 & 485 & 152\\
				Cam 2 & 112,523 & 83 & 94 & 653 & 830 & 152\\
				Cam 3 & 72,005 & 74 & 68 & 822 & 964 & 156\\
				Cam 4 & 53,986 & 70 & 72 & 393 & 536 & 157\\
				Cam 5 & 67,672 & 63 & 52 & 781 & 896 & 158\\
				Cam 6 & 56,472 & 64 & 38 & 865 & 967 & 156\\
				Cam 7 & 17,809 & 48 & 36 & 93 & 177 & 155\\
				Cam 8 & 36,338 & 55 & 40 & 266 & 361 & 149\\
				Cam 9 & 3,910 & 40 & 16 & 32 & 88 & 149\\
				Cam 10 & 10,492 & 62 & 51 & 105 & 218 & 151\\
				Cam 11 & 73,601 & 76 & 149 & 448 & 673 & 145\\
				Cam 12 & 37,543 & 70 & 79 & 233 & 382 & 146\\
				\hline
				Total & 601,581 & 112 & 802 & 3,996 & 6,577 & 1,826\\
				\hline
		\end{tabular}}
		\label{table:stat}
	\end{table}


	In Table \ref{tab:cp_msmc}, we compare RPIfield with existing benchmark multi-shot and multi-camera re-id datasets, noting that the new dataset has the largest number of cameras. In the right column, we summarize the attributes of each dataset as multi-shot (MS), or multi-camera (MC). We emphasize that the probes in our dataset correspond to \textbf{known actors}, who were provided specific walking and re-appearance instructions to aid in the kind of temporal research we discuss in this work. To our knowledge, other re-id datasets were not planned in this way.  While most of the multi-camera or multi-shot re-id datasets have no distractors (except DukeMTMC4ReID \cite{DukeMTMC_CVPR17}, Market-1501 \cite{Market1501_ICCV15} and MARS \cite{MARS_ECCV16}), our dataset preserves all image sequences of all detected distractors. Most importantly, actors could reappear in one or multiple cameras multiple times in our dataset, which allows us to study re-id algorithms in a more general way than usual.

	\noindent
	\begin{table}[!h]
		\centering
		\caption{Comparison of existing multi-shot, multi-camera re-id datasets.}
		\label{tab:cp_msmc}
		\scalebox{0.75}{
			\setlength{\extrarowheight}{.2em}
			\begin{tabular}{|c|c|c|c|c|c|}
				\hline
				\normalsize
				Dataset & \#BBoxes & \#Identities & \#Distractors & \#Cam & Attributes\\
				\hline
				\small
				\textbf{RPIfield} & \textbf{601,581} & \textbf{112} & \textbf{3,996} & \textbf{12} & \textbf{MS, MC} \\
				DukeMTMC4ReID \cite{DukeMTMC_CVPR17} & 4,6261 & 1,852 & 21,551 & 8 & MS, MC\\
				Market-1501 \cite{Market1501_ICCV15} &	2,668 & 1,501 & 2,793 & 6 & MS, MC\\
				MARS \cite{MARS_ECCV16} & 1,067,516 & 1,261 & 3,248 & 6 & MS, MC\\
                SAVIT-Softbio \cite{SAVIT-Softbio_DICTA12} & 64,472 & 152 & 0 & 8 & MS,MC\\
				PRID \cite{PRID_SCIA11} & 2,4541 & 178 & 0 & 2 & MS\\
				iLIDSVID \cite{iLIDS-VID_ECCV14} & 42,495 & 300 & 0 & 2 & MS\\
				3DPeS \cite{3DPeS_ACM11} & 1,011 & 192 & 0 & 8 & MC\\
				WARD \cite{WARD_ECCV08} & 4,786 & 70 & 0 & 3 & MC\\
				CUHK02 \cite{CUHK02_CVPR13} & 7,264 & 1816 & 0 & 10 & MC\\
				CUHK03 \cite{CUHK03_CVPR14} & 13,164 & 1360 & 0 & 10 & MC\\
				RAiD \cite{RAiD_ECCV14} & 6,920 & 43 & 0 & 4 & MC\\
				HDA+ \cite{figueira2014hda} & 2,976 & 74 & 0 & 12 & MC\\
				\hline
		\end{tabular}}
	\end{table}

	\section{Rank Persistence}
	\label{sec:RP}
	In this section, we present the concept of rank persistence in steps, working up to evaluating situations in which multiple persons of interest each reappear one or multiple times within a video. We introduce the \textbf{Rank Persistence Curve (RPC)}, which encapsulates the behavior of correct matches in a temporally evolving and increasing gallery.
	
	For the purposes of illustration, all plots presented in this section are generated by the unsupervised GOG algorithm \cite{GOG_CVPR16} to describe person images, followed by Euclidean distance to rank candidates. We denote this by ``GOG+$l_{2}$" in the plots. These preliminary plots are meant only to motivate the RPC idea, not to propose a new or definitive re-id algorithm.  Further implementation details and performance evaluations are provided in Section~\ref{sec:Exp}.
	
	\subsection{One probe, one reappearance}
	\label{sec:RPC_case1}
\begin{figure*}[h!]
		\centering
        \begin{tabular}{ccc}
\includegraphics[width=0.32\linewidth]{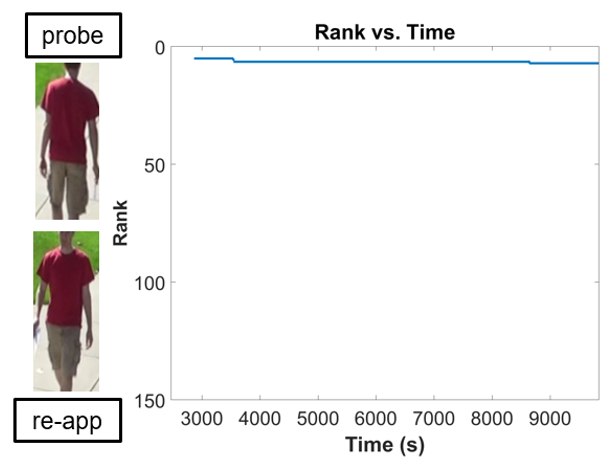} & 
\includegraphics[width=0.32\linewidth]{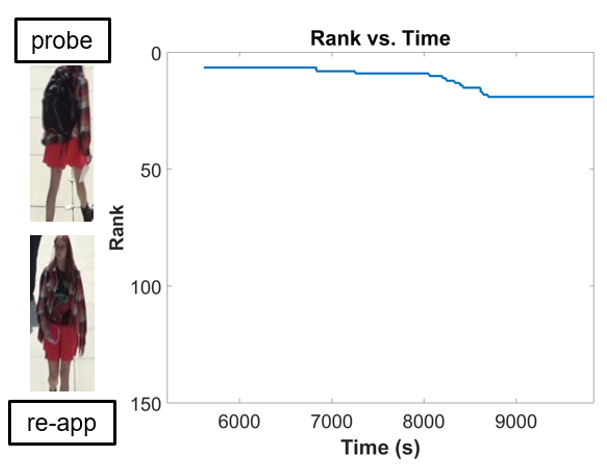} & 
\includegraphics[width=0.32\linewidth]{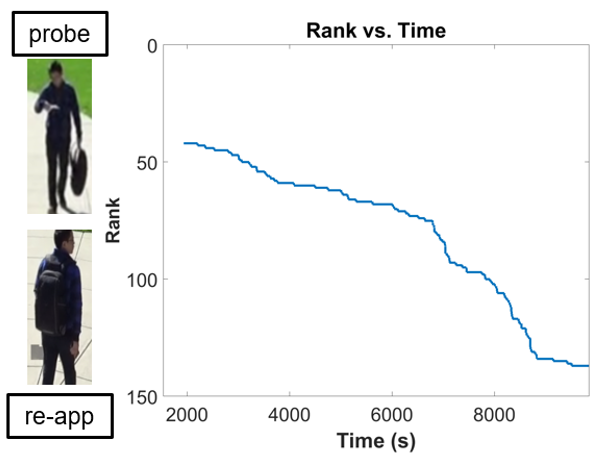} \\ (a) & (b) & (c)
\end{tabular}
		\caption{A comparison of temporal rank curves for 3 different probes that each reappear only once. For each case, the top image is a sample image selected from the probe image sequence, and the bottom image is a sample selected from the known reappearance of the probe in the gallery.} 
		\label{fig:case1_cp_demo}
	\end{figure*}

	We first consider the case of a single probe/query that has exactly one re-appearance over the course of a video sequence. As noted previously, different from existing re-id evaluation schemes where the probe is matched to candidates from a fixed gallery, we consider a temporally evolving gallery set whose dynamically varying size depends on the flow of people over time. In other words, the number of candidates is continuously increasing as more people appear in the video sequence. In this scenario, suppose the person of interest first reappears at time $T$ at rank $k$ computed by a certain re-id algorithm. Let $N_T$ be the size of the gallery at this time (i.e., $N_T$ is the total number of people seen in the gallery camera from time $t=0$ to $t=T$). Given these considerations, the rank of the person of interest can only increase as the gallery size increases after time $t=T$. This notion of instantaneous rank of the person of interest as a function of time in an ever-increasing gallery can be easily captured, as illustrated in Figure \ref{fig:case1_cp_demo}.
	
	
	In each example of Figure \ref{fig:case1_cp_demo}, the horizontal axis is real time (seconds from the beginning of the video sequence) and the vertical axis is the instantaneous rank of the person of interest. Each temporal rank curve does not start from $T=0s$, since the person of interest generally re-appears midway through the video. In this example, each plot ends at $T=9840s$, which is the duration of the entire video sequence. For example, in Figure  \ref{fig:case1_cp_demo}(c), the person of interest first re-appears at rank $k=42$, then drops constantly from $T=1939s$ to the end of the video. This rate of drop in rank basically reflects the robustness of the re-id algorithm to the temporal aspect of matching a probe candidate to an ever-expanding gallery.
	
	Figure \ref{fig:case1_cp_demo} compares these temporal rank curves for three different probes with respect to the same gallery, where each of the probes re-appears only once in the entire video. For the first probe, the rank persists at a relatively low value from the time of its first re-appearance, mostly due to strong appearance similarity between the re-appearance image and the original probe image. In the second case, the re-appearance image is slightly different from the probe image, which results in a steeper drop in the instantaneous rank. Finally, in the third case, the difference between the re-appearance and the probe image is quite obvious, resulting in a more dramatic drop in the instantaneous rank.
    
	\begin{figure*}[h!]
		\centering
        \begin{tabular}{ccc}
\includegraphics[width=0.32\linewidth]{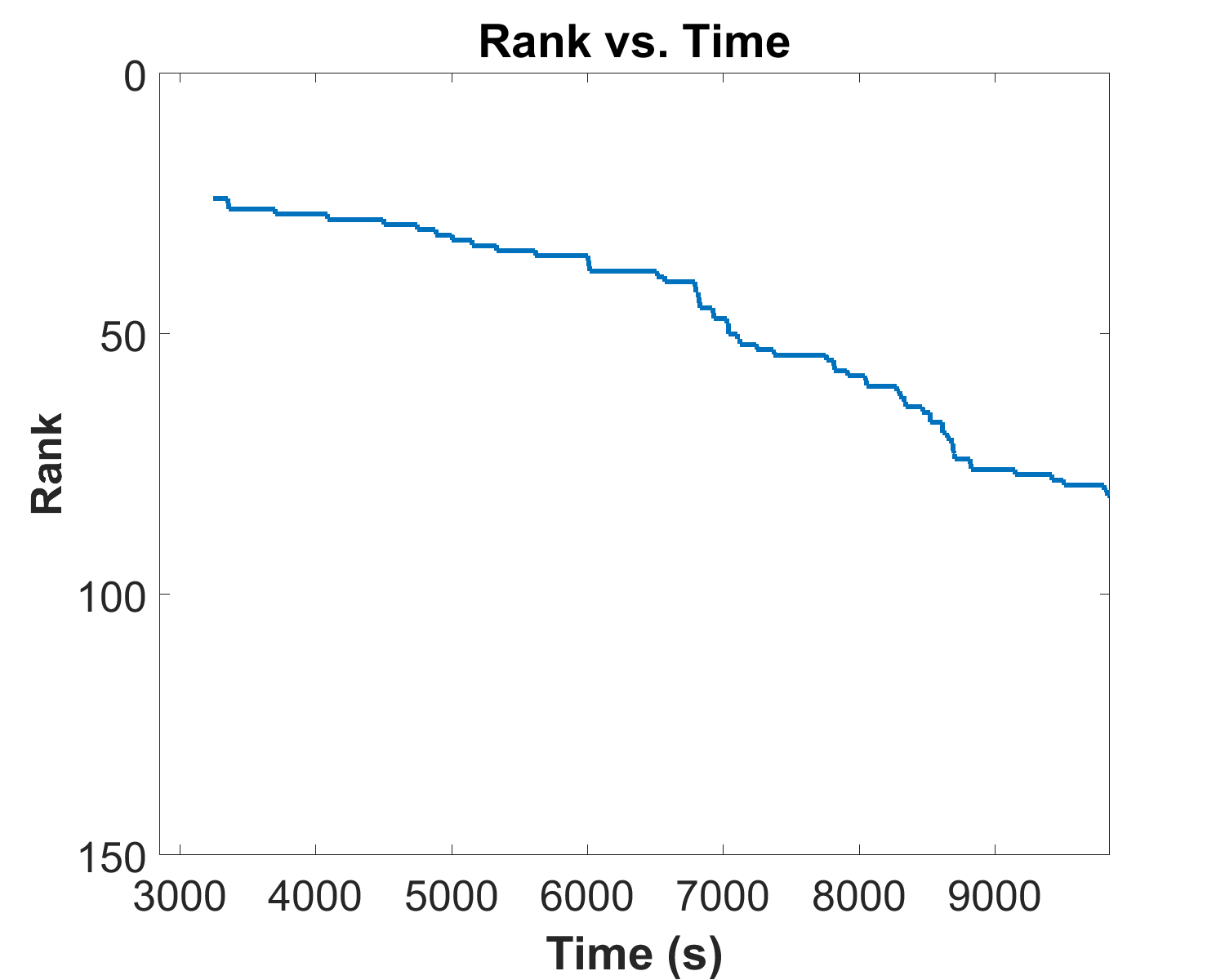} & 
\includegraphics[width=0.32\linewidth]{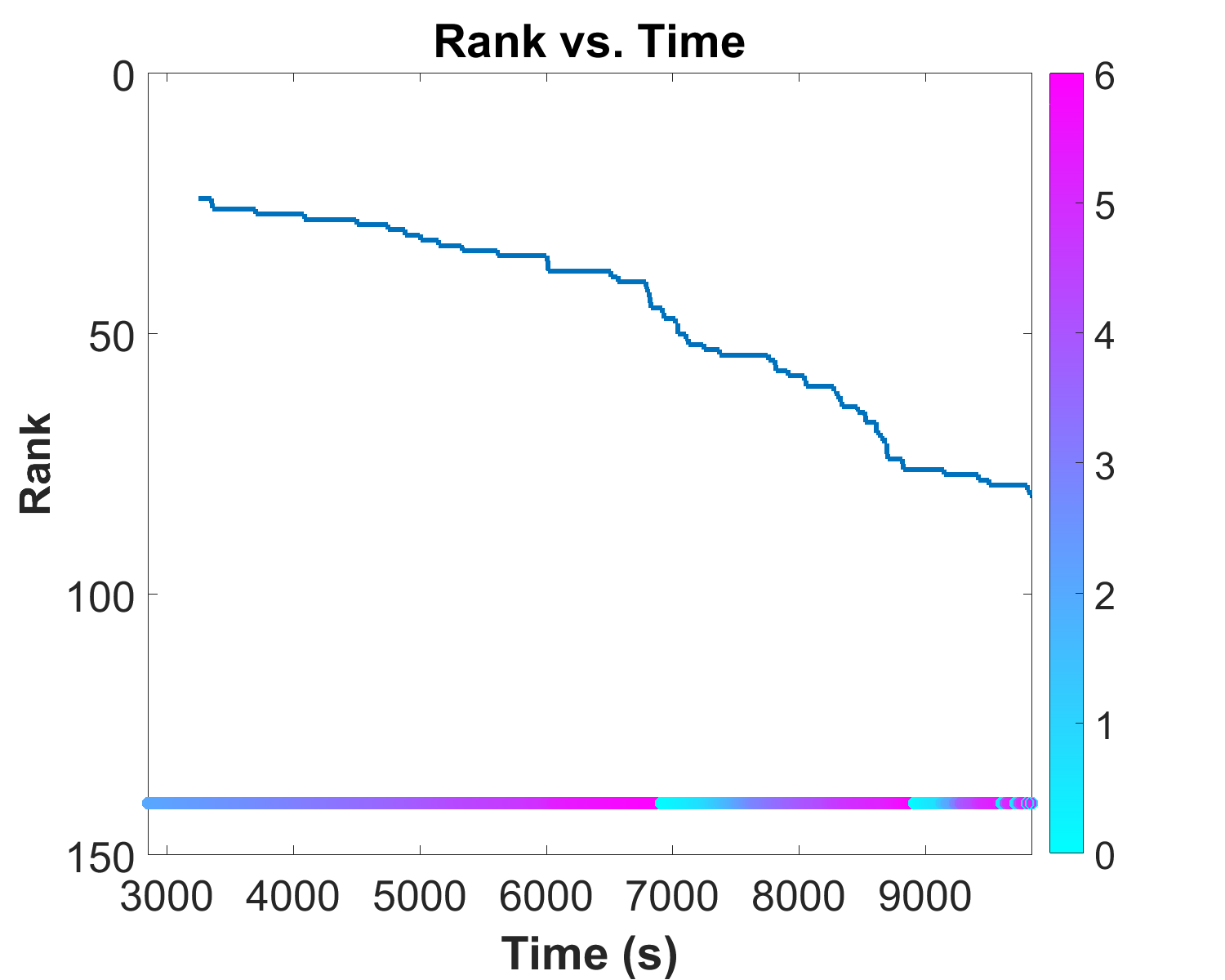} & 
\includegraphics[width=0.32\linewidth]{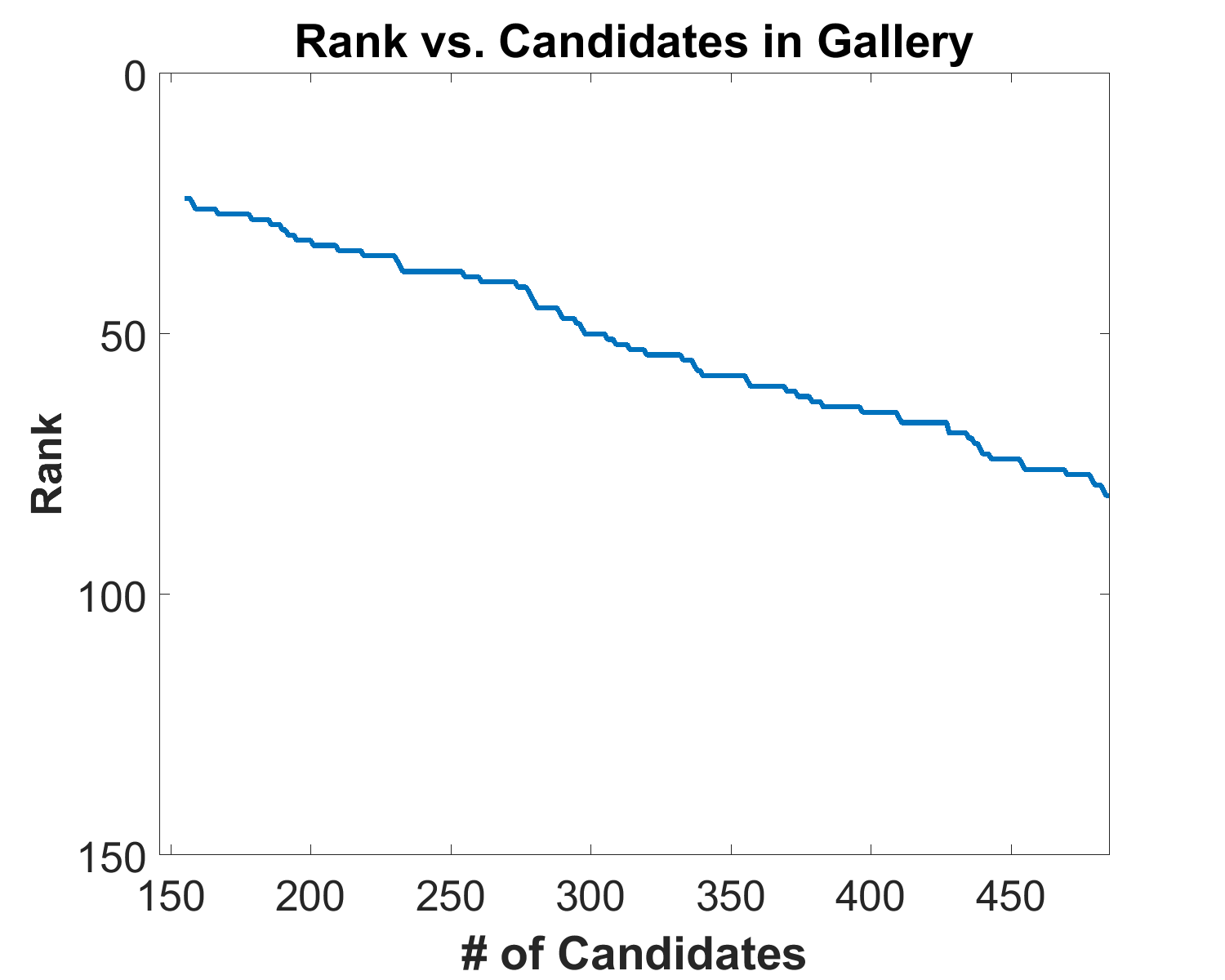} \\ (a) & (b) & (c)
\end{tabular}
\caption{The rank of a person of interest changes in an ever-expanding gallery.  (a) illustrates our usual method for creating temporal rank curves.  (b) is an alternate visualization in which a colorbar indicates the instantaneous video flow density.  (c) is an alternate visualization in which the horizontal axis indicates the number of candidates, not the number of seconds.}
		\label{fig:case1_vfd_demo}
	\end{figure*}
	
	
	One aspect that has a direct impact on these temporal rank curves is the video flow density, defined as the instantaneous number of people seen by the gallery camera per unit time. Specifically, if $N$ people appear in the time range $t=T$ to $t=T+t_0$ in the video sequence, we calculate the video flow density $V$ for this range as:
	\begin{equation}
	V_{(T:T+t_0)} = \frac{N}{t_0}
	\label{eq:vfd}
	\end{equation}
	To understand how this might impact rank persistence, consider two non-overlapping time blocks with the same duration $t_{0}$. In the first block, the gallery camera sees $x$ people walking by. In the second block, the gallery camera sees $y$ people walking by. If $x \gg y$, we add many more people to the gallery in the first time block, likely resulting in a steeper drop in the probe's rank compared to the second time block. 
	
	We considered two ways to visualize and understand the influence of video flow density on our temporal rank curves, shown in Figures \ref{fig:case1_vfd_demo}(b)-(c). Figure \ref{fig:case1_vfd_demo}(b) shows the temporal rank curve with a color bar at the bottom indicating the instantaneous video flow density (the unit time $t_0$ in \ref{eq:vfd} is chosen as $10$ seconds). The purple color in the color bar indicates a large video flow density, whereas blue indicates a lower video flow density. In line with what we discussed above, the high video flow densities at times $t \approx 6500s, 8600s$ cause steep drops in the rank of the probe. 
    
    Figure \ref{fig:case1_vfd_demo}(c) shows an alternate way of visualizing this aspect; the rank of the probe is plotted against the number of gallery candidates. Here, the person of interest first re-appears at rank 24 when the number of candidates in the gallery is 155. Subsequently, the rank curve drops as the number of candidates in the gallery increases. Though video flow density has a direct influence on the per-probe temporal rank curves shown in Figure \ref{fig:case1_cp_demo}, its definition is based on absolute real time. Since the arrival time of each probe belongs to a different time block, each per-probe plot goes through different time periods with various video flow densities. However, the representations in Figures \ref{fig:case1_vfd_demo}(b) and (c) would pose significant difficulty when we  integrate these curves across multiple probes in Section \ref{sec:RPC_case3}. Thus, going forward we only consider real time as the horizontal axis for temporal rank curves. 
	
	\subsection{One probe, multiple reappearances}
	\label{sec:RPC_case2}
Generally, a person of interest might re-appear multiple times during the entire course of a video (i.e., separate reappearances spaced apart in time, not different samples from a tracklet as in multishot re-id) . To plot a temporal rank curve as in Section~\ref{sec:RPC_case1}, we modify the vertical axis to show the lowest instantaneous rank of \textit{any} re-appearance of a probe. 
	
	As in the first case, the instantaneous rank of the probe can only increase after the first re-appearance as more candidates are added to the gallery. However, the rank of the second re-appearance might achieve a lower value than the first re-appearance depending on its similarity to the probe image. After the second re-appearance, the rank curve will again be non-decreasing until we hit the third re-appearance, and so on. Thus, the temporal rank curve is piecewise monotonically non-decreasing. 
    
\begin{figure*}[h!]
		\centering
		\includegraphics[width=.9\linewidth]{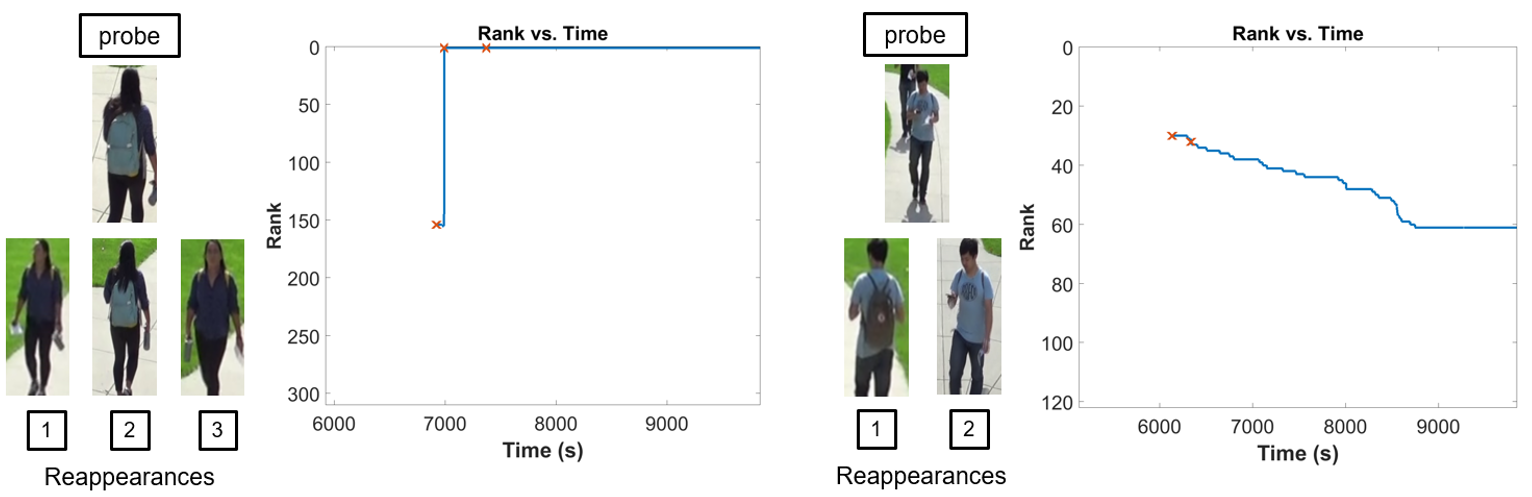}%
		\caption{A comparison of temporal rank curves for 2 different probes that reappear  multiple times. For each case, the top image is a sample image selected from the probe image sequence, and the bottom images are samples selected from the separate reappearances of the probe in the gallery. Each red `$\cross$'  indicates one reappearance of the probe.}
		\label{fig:case2_cp_demo}
	\end{figure*}

    Figure \ref{fig:case2_cp_demo} shows temporal rank curves for two different probes, each of which re-appears multiple times in the same video. For the first probe in Figure \ref{fig:case2_cp_demo}, the curve starts from a large rank (154) at the time of her first re-appearance, since the front and back views differ substantially from each other. At the time of her second re-appearance ($T=6994s$), the rank decreases to 1 since the second re-appearance is very similar to the probe image. This low rank dominates the third re-appearance of the probe. For the second example, the rank curve starts at rank 30 and keeps increasing after the first reappearance. The second reappearance of the probe has no obvious influence on the overall rank curve, despite its higher similarity to the probe image.
	
	\begin{figure*}[h!]
		\centering
        \begin{tabular}{cc}
\includegraphics[width=0.4\linewidth]{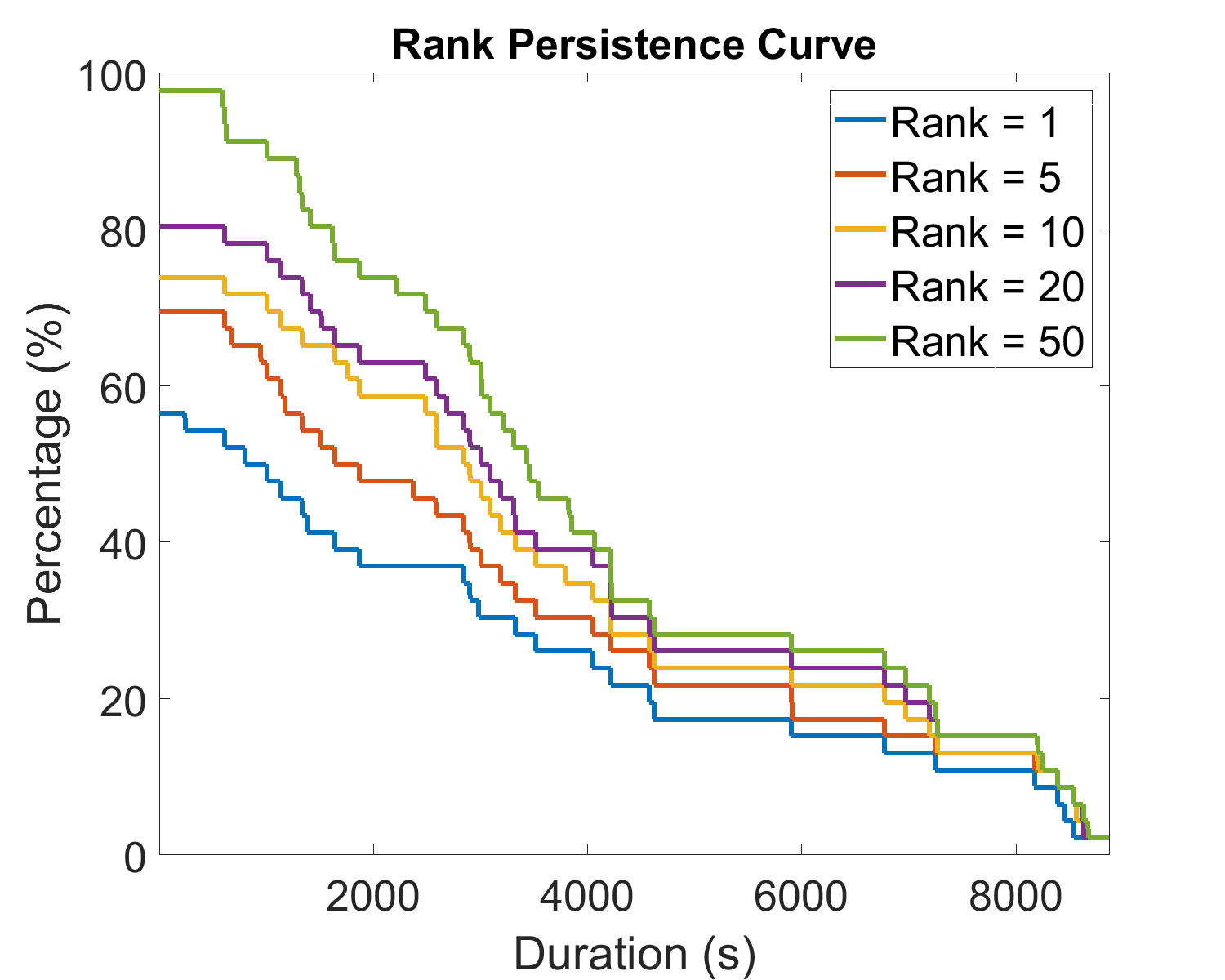} &
\includegraphics[width=0.4\linewidth]{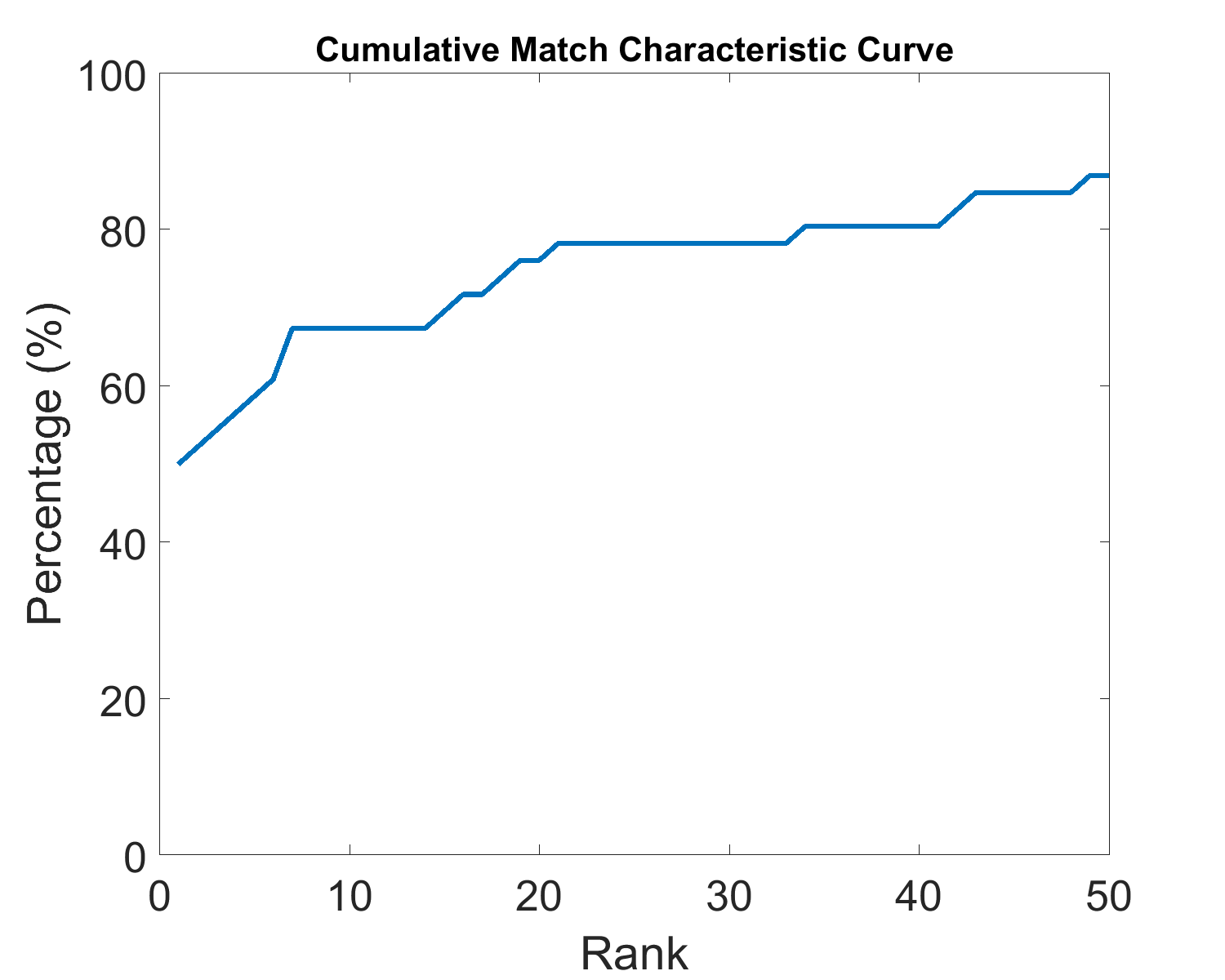} \\ (a) & (b) 
\end{tabular}
		\caption{Comparison of (a) the proposed Rank Persistence Curve (RPC) and (b) the traditional Cumulative Match Characteristic (CMC) curve for Camera 1 of RPIfield, containing 46 probes.}
		\label{fig:case3_demo}
	\end{figure*}
    

	\subsection{Multiple probes, multiple reappearances}
	\label{sec:RPC_case3}
	Finally we arrive at the most general situation, in which we evaluate the performance of a given re-id algorithm across multiple probes, each of which may reappear one or multiple times over the entire course of the video. In this case, we define the \textbf{Rank Persistence Curve} based on the per-probe temporal rank curves introduced in Section \ref{sec:RPC_case1} and \ref{sec:RPC_case2}. The RPC as presented here can be used to evaluate the performance of algorithms on entire datasets that have data structured and time-stamped as described in this work. In the following, we also present a theoretical and empirical comparison with the traditionally used CMC curve to highlight the differences and advantages of the proposed RPC.
	
	For each probe that has at least one re-appearance in the video sequence, we plot the temporal rank curve as described in Sections~\ref{sec:RPC_case1} and~\ref{sec:RPC_case2}. We then define the Rank Persistence Curve (RPC) as follows: The dependent axis of the RPC is set as the duration $d$ in real units (e.g., seconds). For a fixed rank $r$ and each duration $d$, we plot the percentage of probes that appear continuously at or below rank $r$ for at least $d$ units. Based on the definition, the RPC for a fixed rank $r$ is monotonically decreasing, and RPCs at higher ranks dominate those at lower ranks.
	
	Figure \ref{fig:case3_demo}(a) shows the RPC across a dataset of 46 probes in the view of Camera 1 of RPIfield. In the plot, we show RPCs for five different ranks $r \in \{1, 5, 10, 20, 50\}$. In contrast, the traditional CMC for the same set of 46 probes and using the same algorithm is shown in Figure \ref{fig:case3_demo}(b). We can see that the two types of curves are qualitatively different. Since we want to capture the temporal aspect of rank in the RPC, the dependent axis is no longer rank but duration, and we need a third ``axis'' (in this case color) to indicate rank.
    
	
To understand the RPC, let us consider the RPC for $r=1$, shown in blue.  This captures our objective of visualizing how likely and for how long a candidate is to stay at rank 1 across a long video sequence. The RPC starts at 57\% and stays there for 241 seconds, meaning 57\% of the probes had a re-appearance at rank 1 that persisted for 241 seconds. The RPC is a monotonically non-increasing curve with respect to the duration, which is intuitive since the lowest instantaneous rank of \textit{any} re-appearance of a probe can either stay same or increase as we add more candidates over time to the gallery. Of course, rank 1 is a stringent requirement, and candidates are more likely to persist for longer durations at higher ranks. This is evident from the RPCs for $r>1$  in the plot, where we see much higher percentages of probes that stay at $r>1$ for a given duration $d$.
	\begin{figure*}[!b]
		\centering
		\hspace{-8mm}
        
		\subfloat[Camera 1, Person 5]{\includegraphics[width=0.26\linewidth]{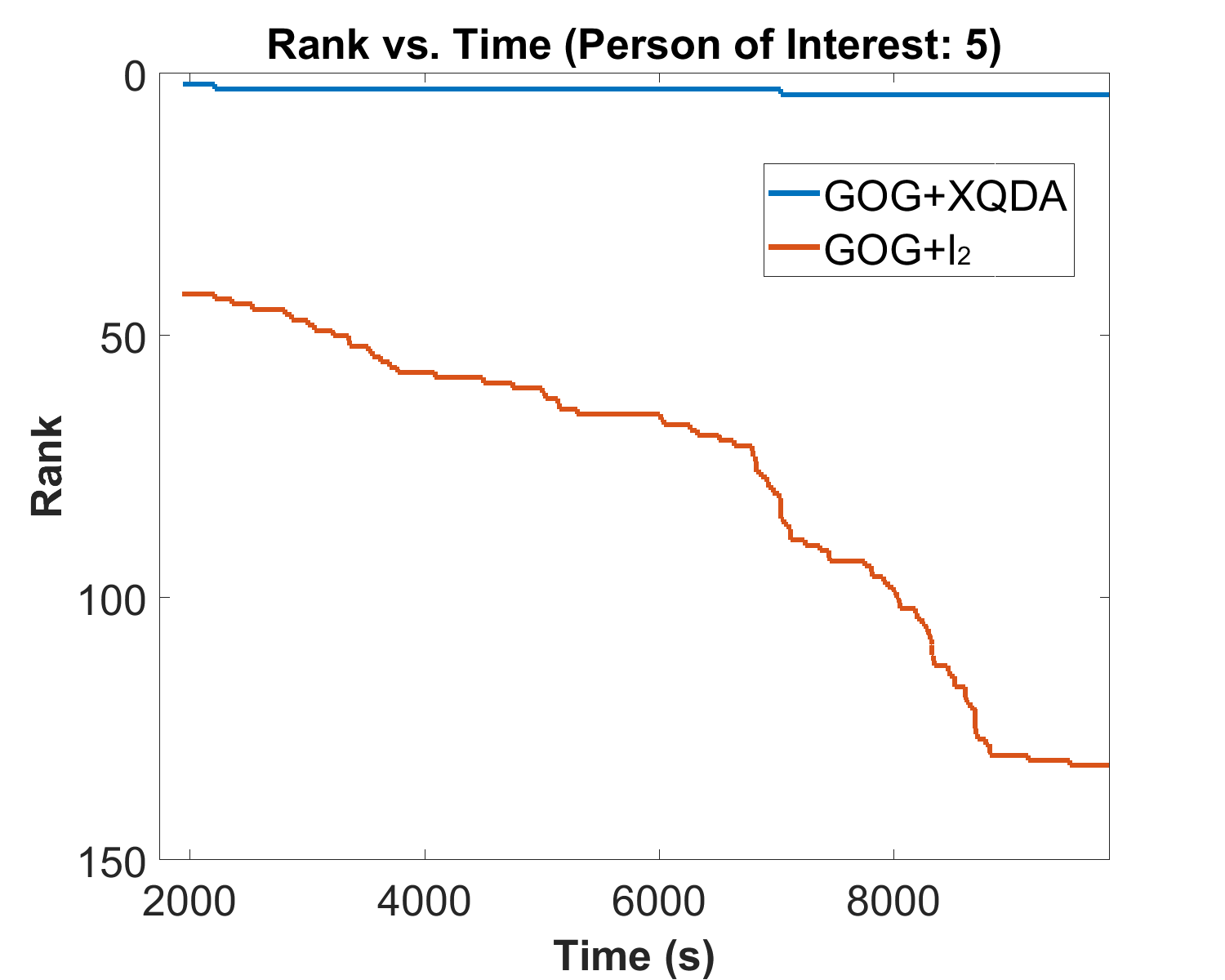}}\hspace{-5mm}
		\subfloat[Camera 1, Person 69]{\includegraphics[width=0.26\linewidth]{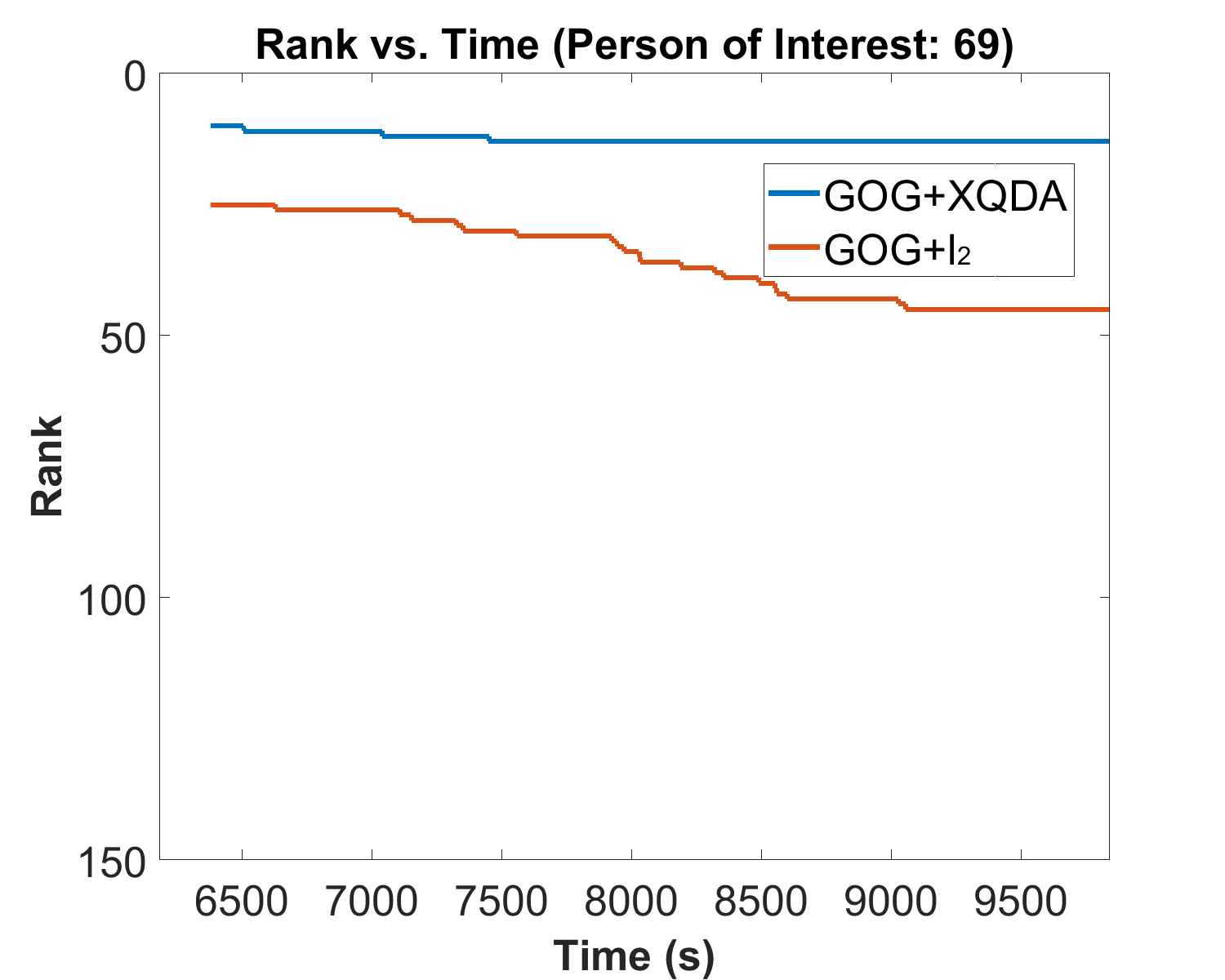}}\hspace{-5mm}
		\subfloat[Camera 4, Person 97]{\includegraphics[width=0.26\linewidth]{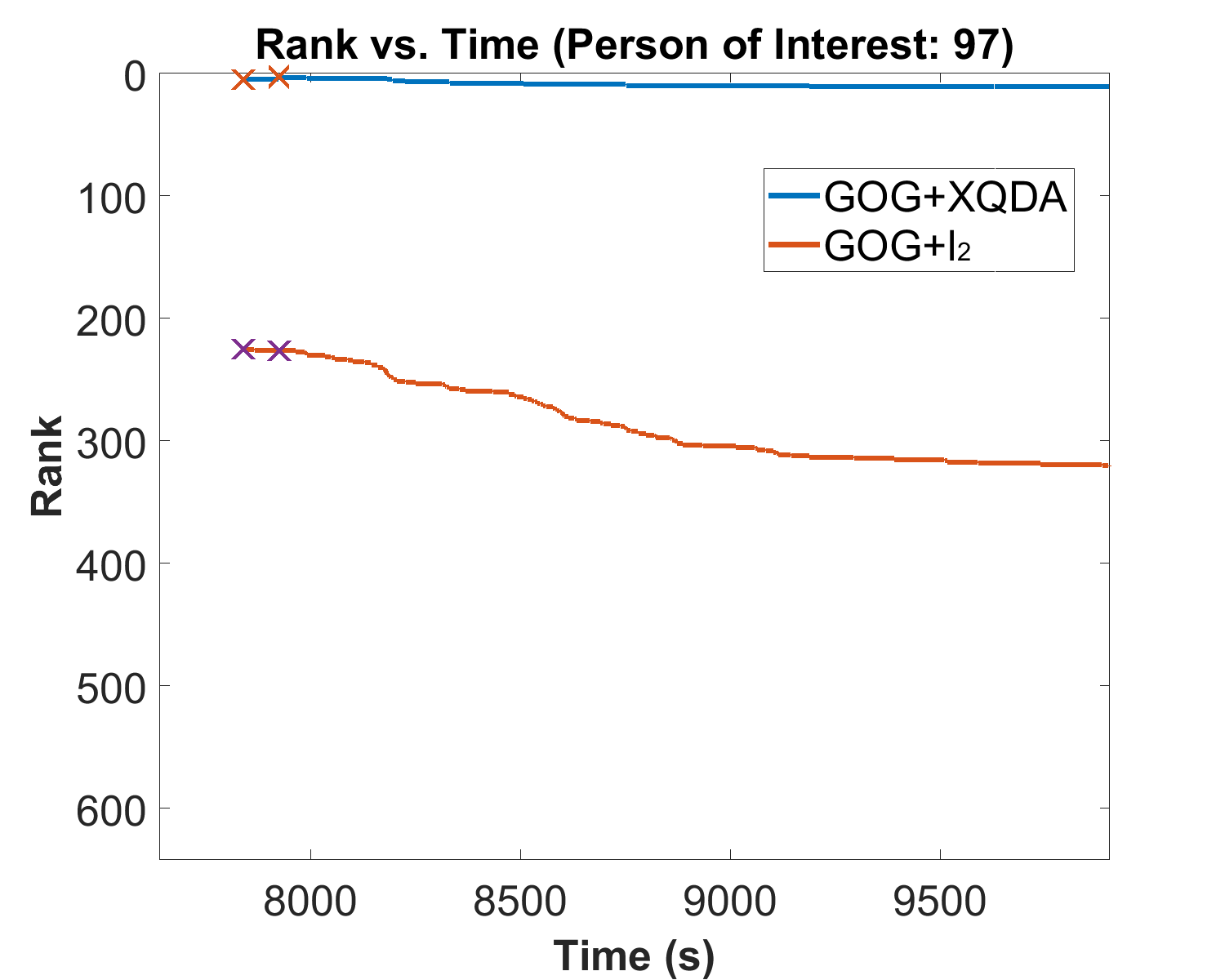}}\hspace{-5mm}
		\subfloat[Camera 4, Person 98]{\includegraphics[width=0.26\linewidth]{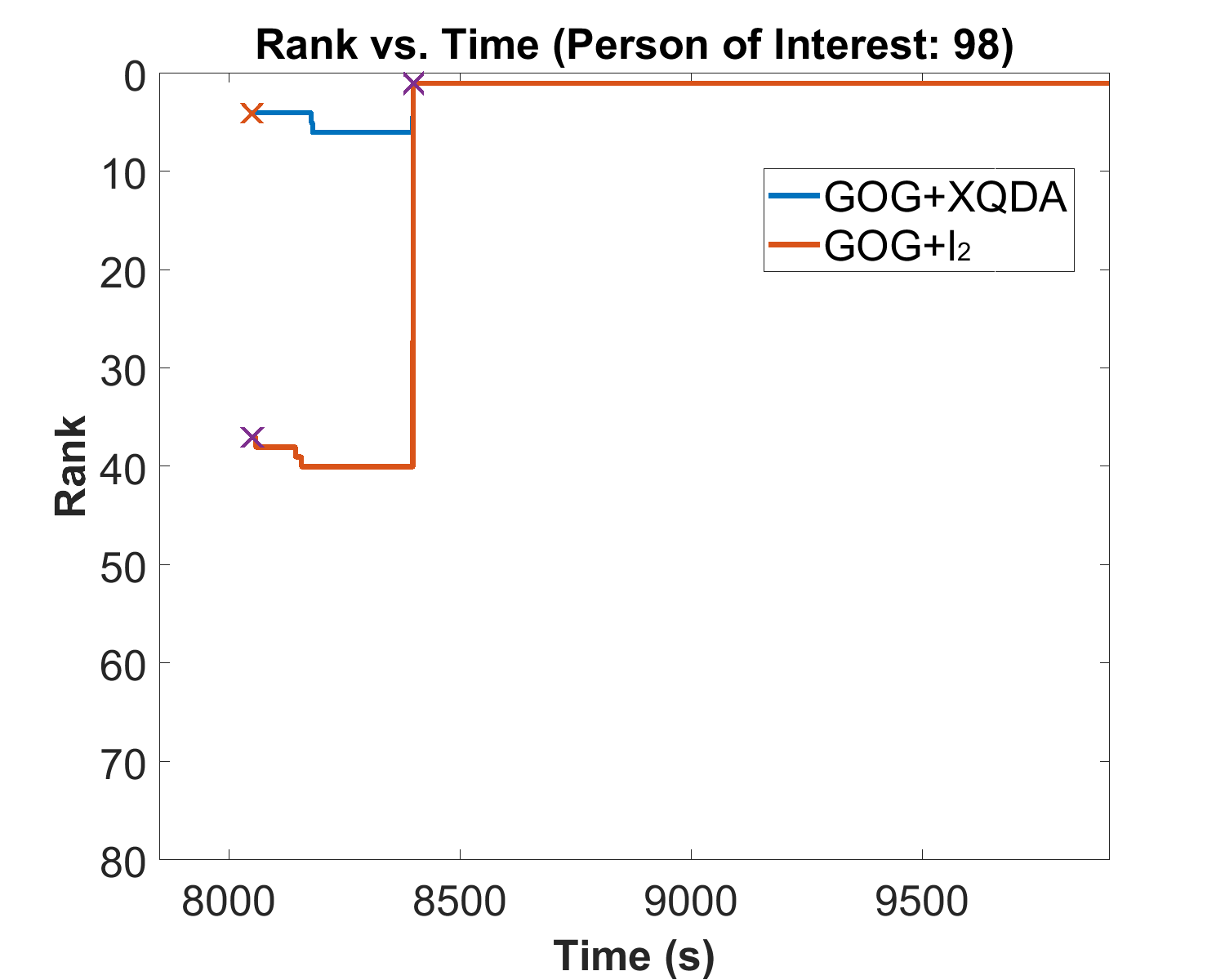}}\hspace{-5mm}
		\caption{Comparison of temporal rank curves of different probes from Cameras 1 and 4. GOG+XQDA (blue curve) uses GOG for feature extraction and XQDA for metric learning; GOG+$l_{2}$ (red curve) uses GOG as feature extraction and $l_2$ distance for ranking (baseline). Panels (a)-(b) show temporal rank curves for participants who reappear once in Camera 1.  Panels (c)-(d) show rank curves for participants who reappear multiple times in Camera 4.} 
		\label{fig:single_cp1}
	\end{figure*}
    
	
	As noted above, the RPC is both qualitatively and measurably different from a CMC curve. Unlike the CMC, the RPC provides valuable information on \textbf{how long} correct matches to the probes persist within the rank-k shortlist in a time-varying gallery. These considerations are important both in terms of the length of the shortlist (real-world end users, typically not computer vision experts, would not want to scroll through pages and pages of candidates to find the person of interest) and the duration of persistence (in real-world scenarios, end users may only get around to checking the output of a re-id surveillance system a few times an hour).  With this theory we can now compare different re-id algorithms and analyze performance more deeply.
	
	\section{Experiments and Results on RPIfield}
	\label{sec:Exp}
	In this section, based on our discussion in Section~\ref{sec:RP}, we quantify the temporal performance of several re-id algorithms using RPCs. We first present single-camera and pairwise-camera results on the RPIfield dataset using a specific re-id algorithm. Subsequently, we compare the performance of various re-id algorithms to demonstrate the suitability of RPCs to evaluate and contrast their temporal performance.  Again, our goal in this paper is not to promote a particular re-id algorithm, but a method for algorithm performance evaluation.
	
	We first briefly describe the basic re-id algorithm for our initial experiments. We use the recently proposed Gaussian of Gaussian (GOG) descriptor \cite{GOG_CVPR16} to extract features from person images. Given an image sequence containing $n$ images $I_1,I_2,...,I_n$ for each person, we compute a single feature vector describing the sequence as the average of the $n$ feature vectors $\bm{f}_1,\bm{f}_2,...,\bm{f}_n$ computed with GOG.
	
	We present re-id results using the Euclidean ($l_2$) distance, which we refer to as the ``baseline'' algorithm, as well as the XQDA \cite{LOMO_XQDA_CVPR15} distance metric. To train the XQDA distance metric, we randomly select 20 identities from the 112 in RPIfield and use image sequences from all the 12 camera views.  To form positive training pairs, we take all possible combinations of the tracks of the same person from all camera views. For instance, if a participant appears 3 times in Camera 1, 2 times in Camera 2 and 3 times in Camera 6, we will form 3$\times$2 = 6 different positive pairs for camera pair (1, 2), 3$\times$3 = 9 pairs for camera pair (1, 3), and 2$\times$3 = 6 pairs for camera pair (2, 3) respectively. All these pairs are then used to train the metric. The results presented here are on a testing set comprised of the 92 participant image sequences not used during metric training.
	
	

	\begin{figure*}[h!]
		\centering
		\subfloat[RPC for Camera 1]{\includegraphics[width=0.35\linewidth]{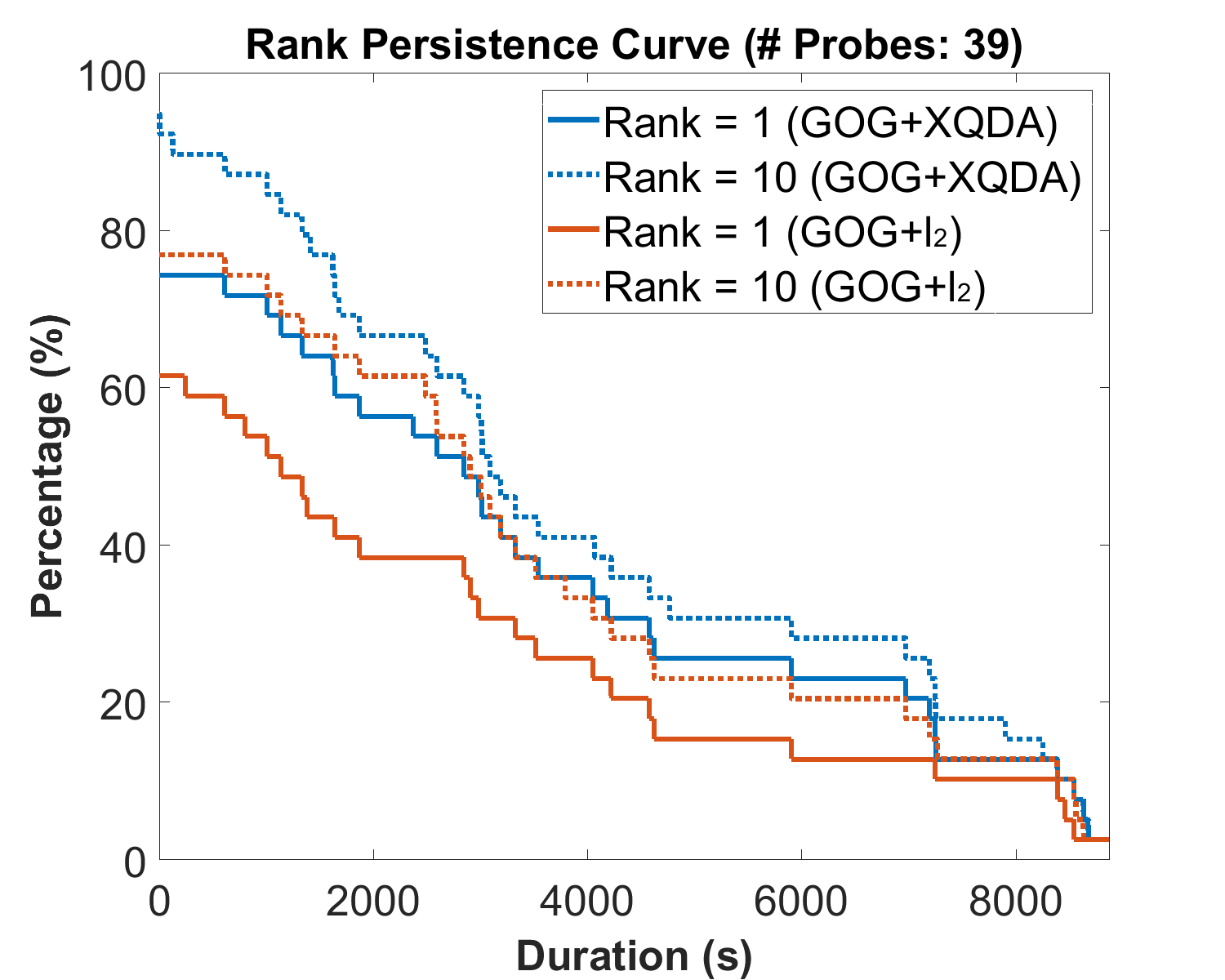}}%
		\qquad
		\subfloat[RPC for Camera 4]{\includegraphics[width=0.35\linewidth]{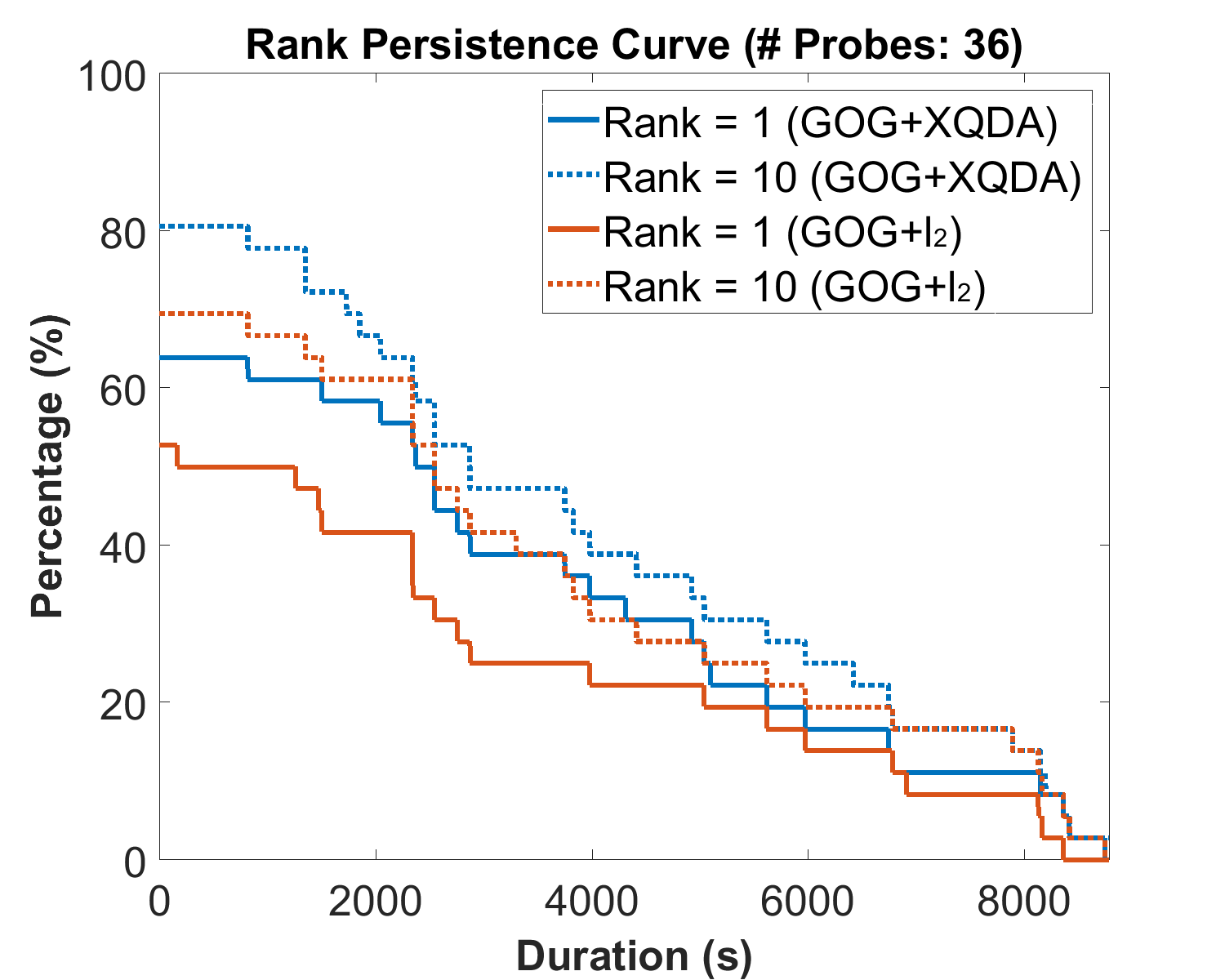}}%
		\\
		\subfloat[CMC curve for Camera 1]{\includegraphics[width=0.35\linewidth]{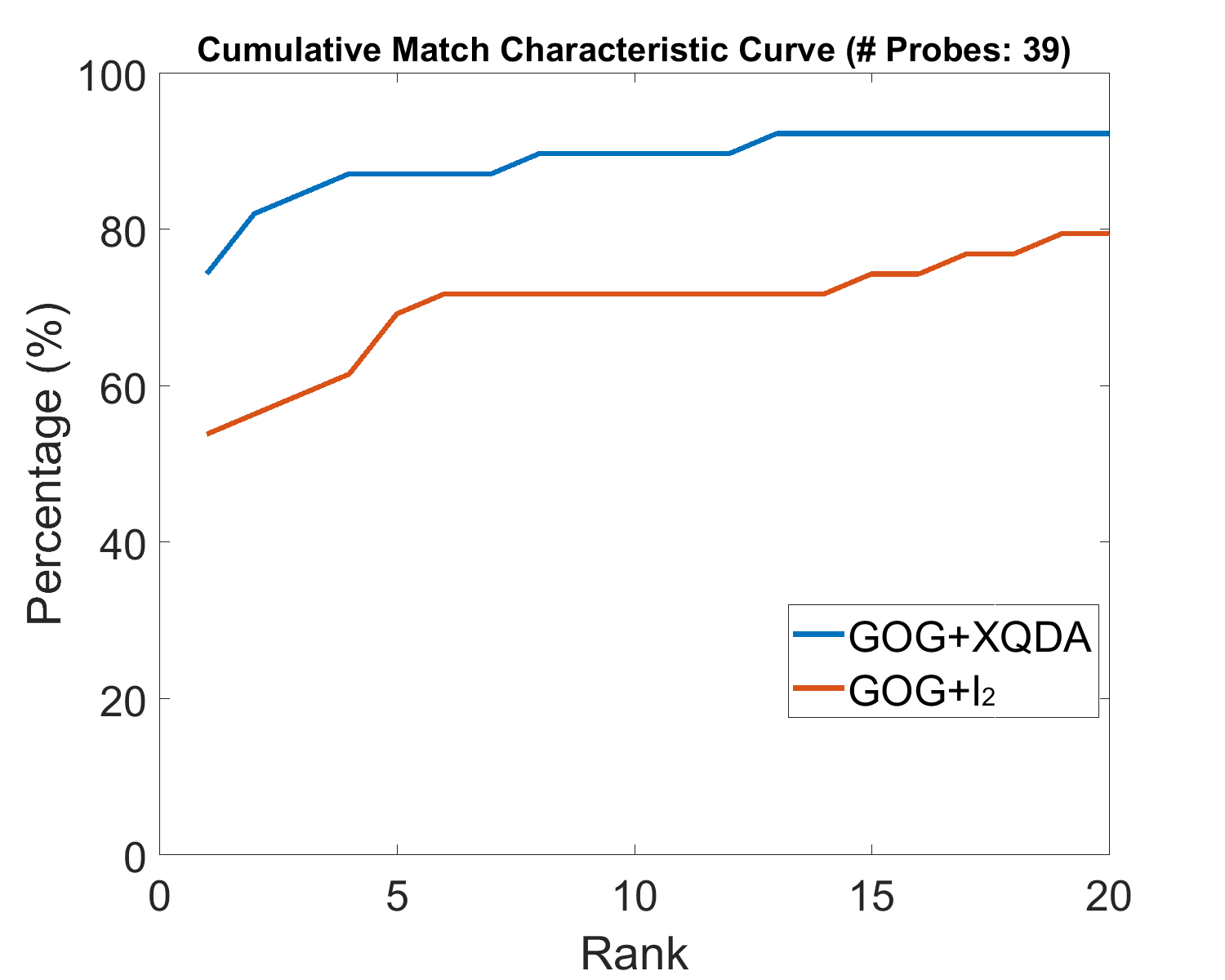}}%
		\qquad
		\subfloat[CMC curve for Camera 4]{\includegraphics[width=0.35\linewidth]{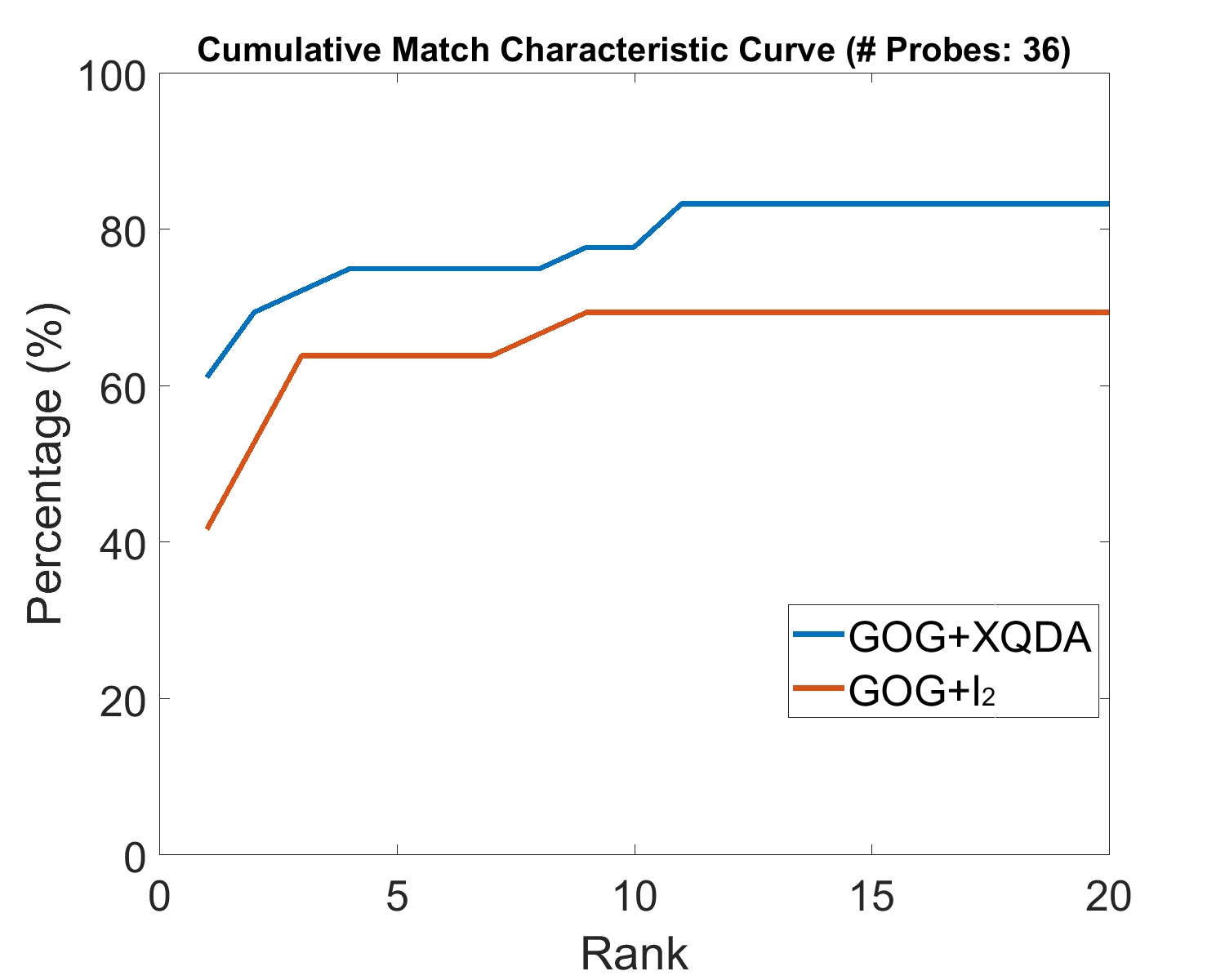}}%
		\caption{(a)-(b) RPCs and (c)-(d) CMC curves for Cameras 1 and 4. ``\# Probes" indicates the total number of probes considered in the gallery. In Panels (a) and (b), GOG+XQDA is compared to the baseline algorithm with fixed ranks $r=1$ and $10$.} 
		\label{fig:single_cp2}
	\end{figure*}

	To plot the temporal rank curve for a probe at any given time $t=T$, we consider all candidates that ever appeared between $t=0$ and $t=T$ as the gallery with which to compare the probe. This assumes at least one re-appearance of the probe in this time duration, which is known from ground truth. In the results that follow, we present the performance of GOG+XQDA (GOG as the feature and XQDA as the distance metric) and compare it with the baseline GOG+$l_{2}$ (GOG as the feature and $l_{2}$ as the distance metric).
	
	\subsection{Single-Camera Temporal Performance Evaluation}
	\label{sec:exp_single}

We begin by performing experiments on single-camera videos from RPIfield, where we use the two algorithms discussed above to analyze temporal characteristics of both the dataset and the algorithms. 
	\subsubsection{Temporal Rank Curves of Single Probes}
	\label{sec:exp_single_one}	
	
    In this section, we present single-probe evaluation results on two best performing camera views (1 and 4). Figure \ref{fig:single_cp1}(a)-(b) shows the temporal rank curves for two participants who each re-appear once in the view of Camera 1. For each rank curve in Figure~\ref{fig:single_cp1}, we compare the instantaneous rank of GOG+XQDA with the baseline GOG+$l_{2}$. Figure \ref{fig:single_cp1}(c)-(d) gives similar plots for participants who re-appear multiple times in Camera 4. Some reappearances of the probes, for instance the second reappearance of Participant 98 (Figure \ref{fig:single_cp1}(d)), cause abrupt decreases in the instantaneous rank, due to their strong similarity to the probe image. 
	
	The varying patterns of the temporal rank curves in Figure~\ref{fig:single_cp1} shed light on several characteristics of the probe and gallery images. For instance, the low drop rate of GOG+XQDA indicates relative temporal robustness of the re-id algorithm, i.e., robustness of the matching result to a temporally evolving and expanding gallery, while the $l_{2}$ metric performs relatively poorly.
	
	\subsubsection{Rank Persistence Curves}
	More generally, multiple probes need to be considered for re-id performance evaluation. As discussed in Section \ref{sec:RPC_case3}, we can incorporate per-probe temporal rank curves into Rank Persistence Curves for single camera re-id, as shown in Figure \ref{fig:single_cp2}. Each RPC represents a particular algorithm's performance on the dataset comprised of the probes considered as part of evaluation.
	
	Figures \ref{fig:single_cp2}(a) and (b) show temporal evaluation results for video from Cameras 1 and 4, while Figures \ref{fig:single_cp2}(c) and (d) show the corresponding CMC curves. For each camera, we aggregate performance over all participants who reappeared at least once during the course of the video. 
	These curves help illustrate important insights from the RPC that the CMC curve ignores. For instance, consider the rank-10 RPC of GOG+XQDA (dotted blue) in Figure \ref{fig:single_cp2}(a). The RPC tells us that the performance drops from 95\% to 3\% across the the entire duration of the video. This means only 3\% of the probes in our dataset ``persist'' at rank 10 when we consider the duration of the entire video. On the other hand, the CMC curve for GOG+XQDA in Figure \ref{fig:single_cp2}(c) only tells us that the rank-10 performance was 90\%, meaning 90\% of the probes were re-identified within rank-10 in a gallery of candidates all considered at the same time; all temporal information is lost.  The RPCs give us more detailed temporal performance evaluation while being easy to present and interpret.
	
\begin{figure*}[htbp!]
		\centering
		\includegraphics[width=0.9\linewidth]{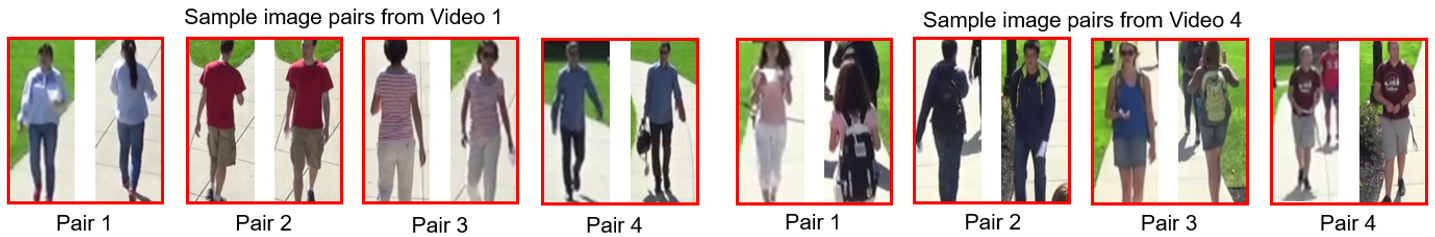}
		\caption{Four example image pairs from Cameras 1 and 4. For each pair, the left is the probe image, and the right is the reappearance image.}
		\label{fig:vis_single_cp}
	\end{figure*}

	Comparing RPCs of the same color and linestyle in Figure \ref{fig:single_cp2}(a) and (b), we can see that both the GOG+XQDA and baseline algorithms generally perform better in Camera 1 than in Camera 4. As illustrated in Figure \ref{fig:vis_single_cp}, this is likely caused by a higher fraction of difficult examples (e.g., participants wearing backpacks), along with a higher number of distractors and more shadow/illumination issues.

	\begin{figure*}[!h]
		\centering
		\subfloat[Person 1]{\includegraphics[width=0.26\linewidth]{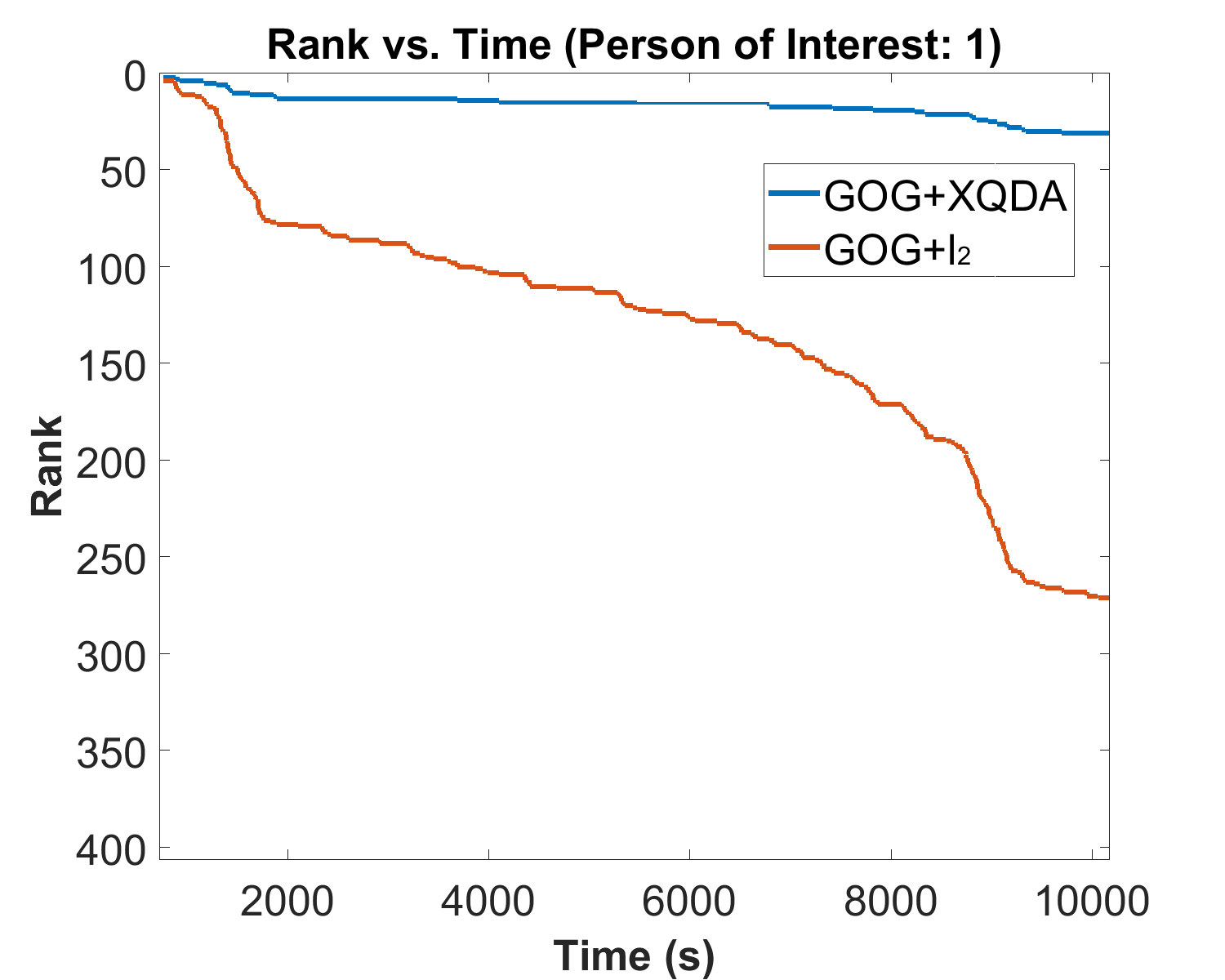}}\hspace{-5mm}
		\subfloat[Person 110]{\includegraphics[width=0.26\linewidth]{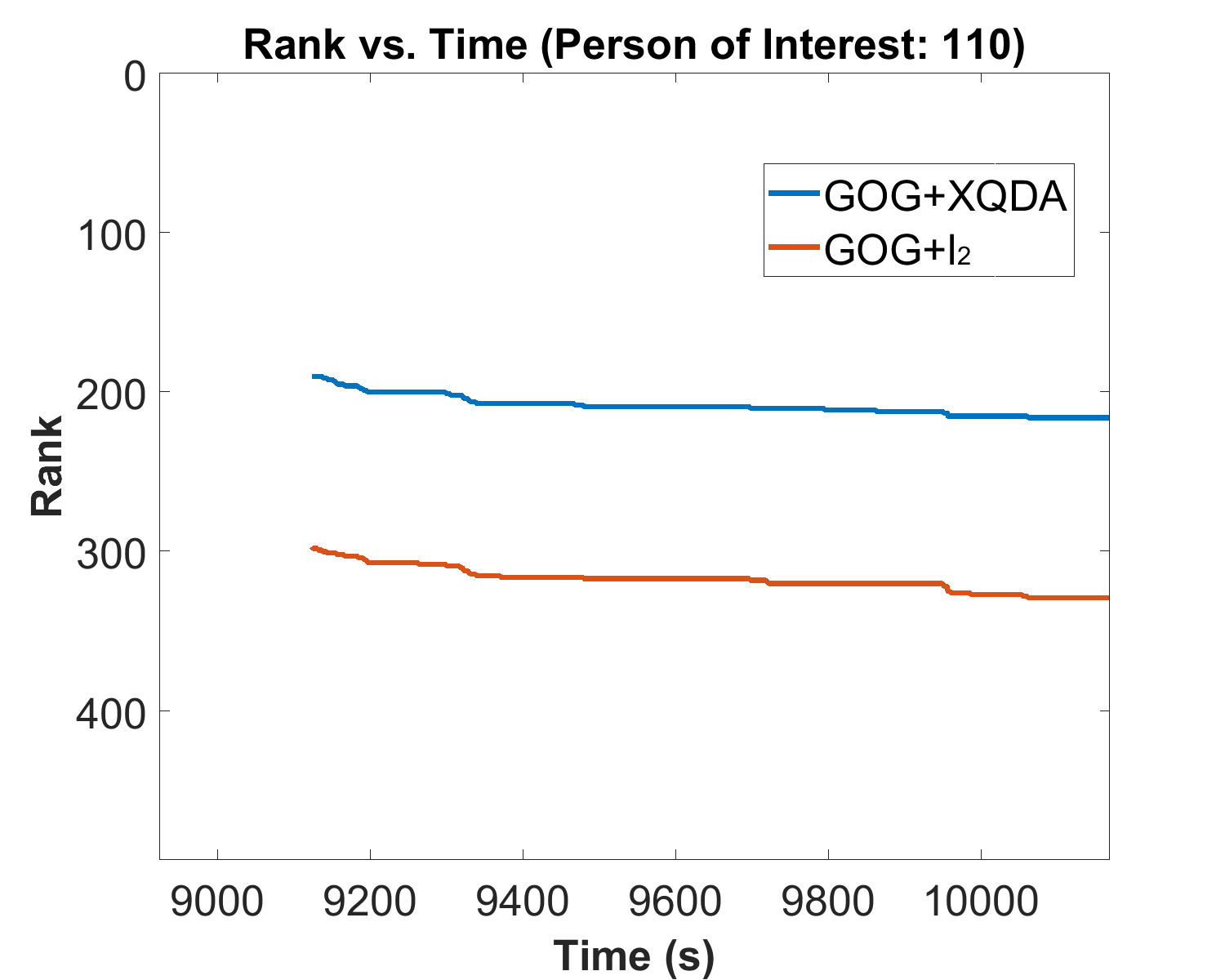}}\hspace{-5mm}
		\subfloat[Person 10]{\includegraphics[width=0.26\linewidth]{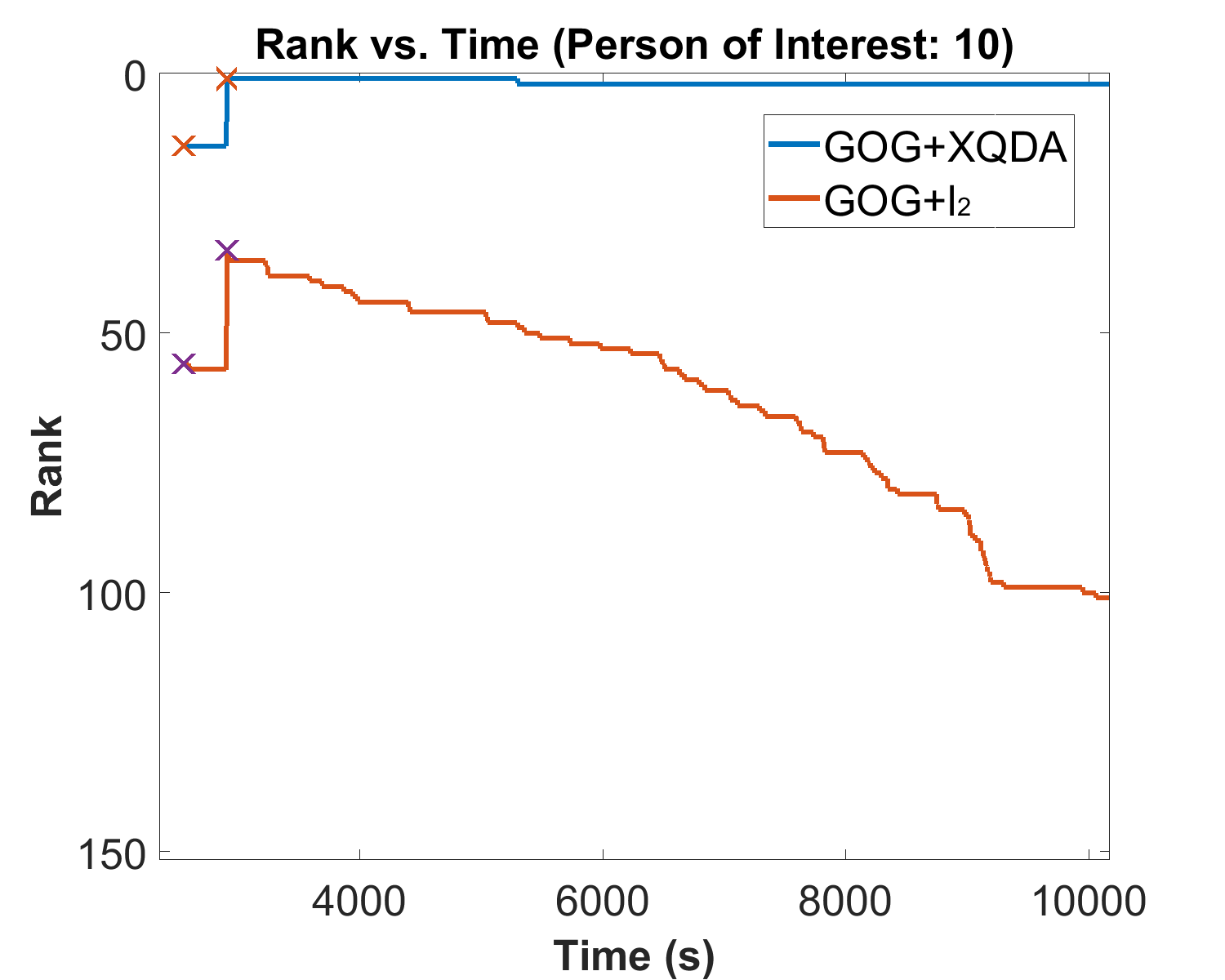}}\hspace{-5mm}
		\subfloat[Person 55]{\includegraphics[width=0.26\linewidth]{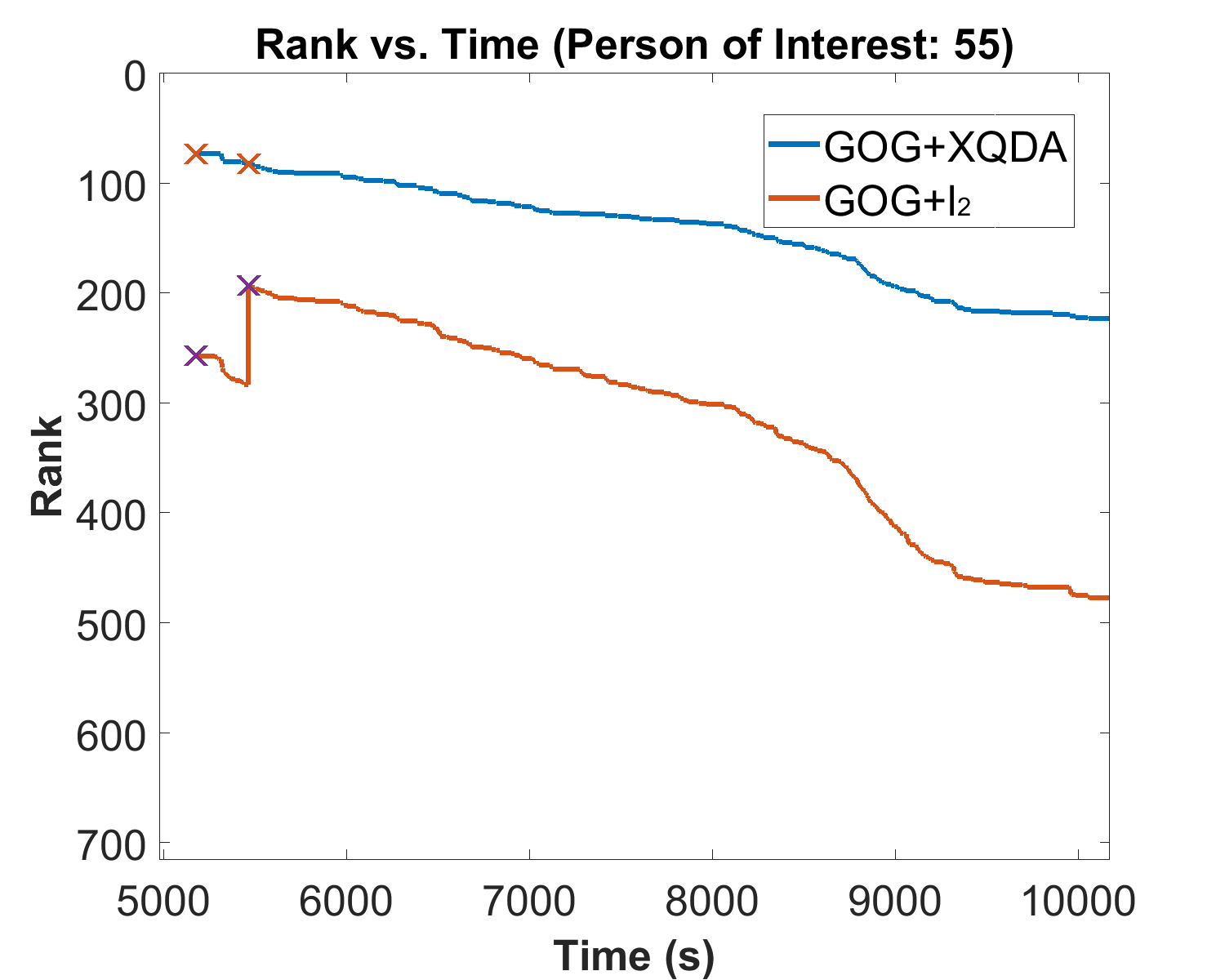}}\hspace{-5mm}
		\caption{Comparison of temporal rank curves of different probes for camera pair (1, 2). Probes are taken from Camera 1, candidates are from Camera 2. Panels (a)-(b) show rank curves for participants who reappear once in Camera 2; Panels (c)-(d) show rank curves for participants who reappear multiple times in Camera 2.}
		\label{fig:pair_cp1}
	\end{figure*}
	
	\begin{figure*}[h!]
		\centering
		\subfloat[Rank Persistence Curves]{\includegraphics[width=0.35\linewidth]{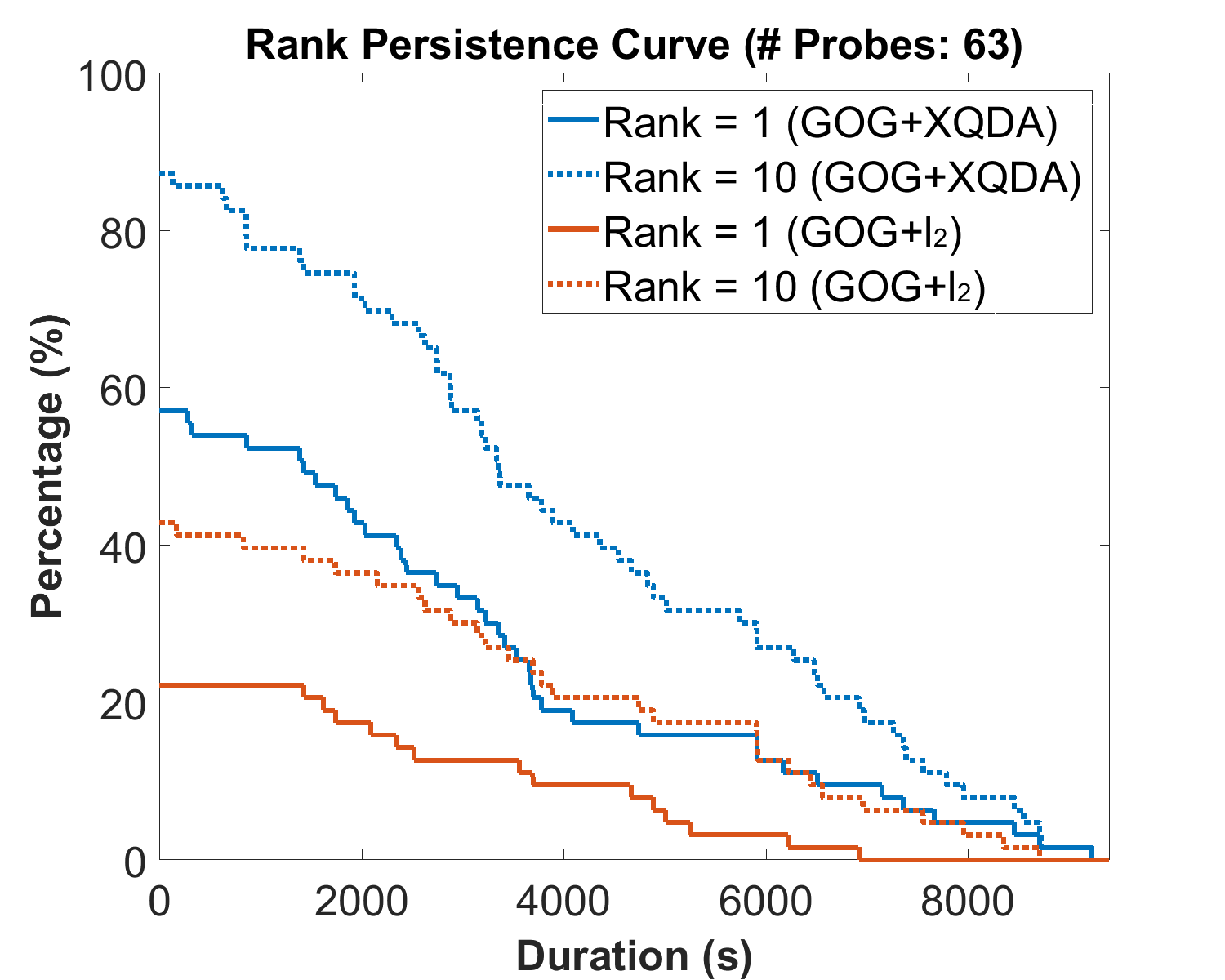}}%
		\subfloat[Cumulative Match Characteristic Curves]{\includegraphics[width=0.35\linewidth]{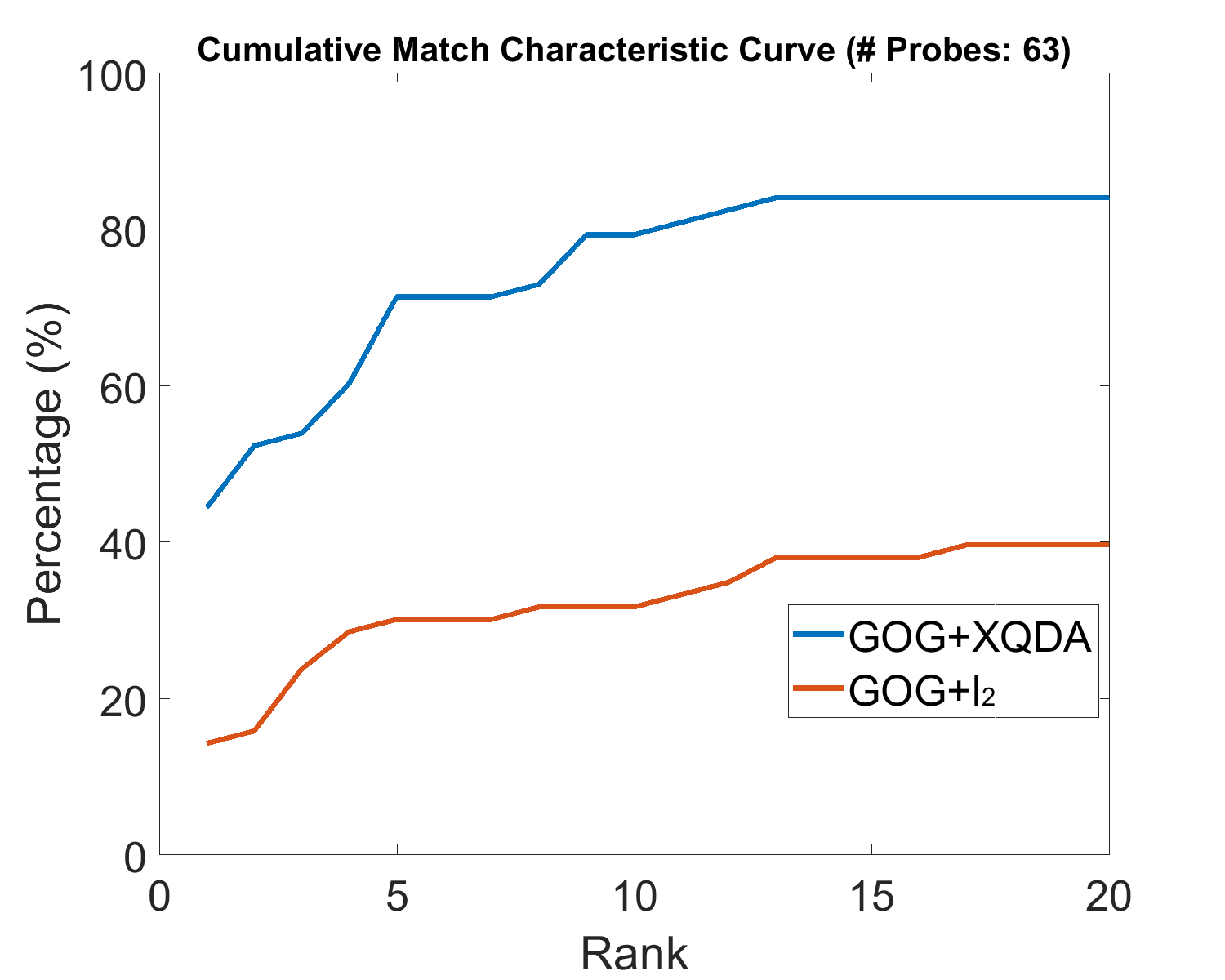}}
		\caption{RPCs and CMC curves of multiple probes for camera pair (1, 2). For RPCs, GOG+XQDA is compared to the baseline algorithm with fixed ranks $r=1$ and $10$. ``\# Probes'' indicates the total number of probes considered in the RPC plot.}
		\label{fig:pair_cp2}
	\end{figure*}

	\subsection{Pairwise-Camera Temporal Performance Evaluation}
	In Section \ref{sec:exp_single}, we evaluated single camera videos from RPIfield using RPCs. In more realistic situations, we need to consider the case that probe and gallery candidates are from different camera views. In cross-view re-id, illumination conditions and viewpoint variations will significantly influence the performance of re-id algorithms, which will also be reflected in our temporal evaluation. In this section, we evaluate and compare the performance of re-id algorithms using RPCs for different camera pairs.
	
	\subsubsection{Temporal Rank Curves of Single Probes}
	\label{sec:exp_pair_one}
	For a person of interest who appears in two different camera views at different times, we consider the image sequence of the person of interest in one of the camera views as the probe and plot the temporal rank curve in the other camera view's time-varying gallery, an example of which is shown in Figure \ref{fig:pair_cp1}. 
	Similar to the results in Section \ref{sec:exp_single_one}, these per-probe rank curves show different patterns. Focusing on the GOG+XQDA (blue) curve in Figures \ref{fig:pair_cp1}(a) and (c), the rank is at a fairly low value due to the strong similarity between the feature vectors of the probe in the projected feature space. In Figures \ref{fig:pair_cp1}(b) and (d), the gallery reappearance/s is dissimilar to the probe, so the rank is relatively high and increases correspondingly as new candidates appear.

	\subsubsection{Rank Persistence Curves}
	We now move to the general case of multiple probes for pairwise re-id evaluation. Using the per-probe rank curves from above, we plot the RPCs for this experiment, shown in Figure \ref{fig:pair_cp2}. For the RPC in Figure \ref{fig:pair_cp2}(a), we consider all participants (63 in total) who appear in Cameras 1 and 2. The corresponding CMC curves for this experiment are shown in Figure \ref{fig:pair_cp2}(b). When compared to the RPCs in Figure~\ref{fig:single_cp2} for single camera views, we can see a larger performance difference between GOG+XQDA (blue lines) and the baseline (red lines) algorithm in Figure \ref{fig:pair_cp2} for pairwise re-id. This indicates that XQDA learning improves cross-view re-id more significantly than single view re-id. We notice a similar trend between the CMC curves in Figure \ref{fig:pair_cp2}(b) and Figures \ref{fig:single_cp2}(c) and (d). From the rank-1 RPCs in Figure \ref{fig:pair_cp2}(a), however, we observe the percentage difference between GOG+XQDA and the baseline curve drops from 35\% to 0\% when we consider the entire duration of the video, while the CMC curves only tell us that the rank-1 performance difference between GOG+XQDA and the baseline is 30\% when candidates in the gallery are all considered at the same time. From the RPCs, we know that even though GOG+XQDA performs much better than the baseline algorithm within the short duration after a probe's reappearance, this performance advantage is continuously decreasing as more candidates are added to the gallery. This information would be important to users of real-world re-id systems when comparing various re-id algorithms.
	
\begin{figure*}[h!]
		\centering
		\subfloat[Probe camera 1, Gallery camera 5]{\includegraphics[width=0.35\linewidth]{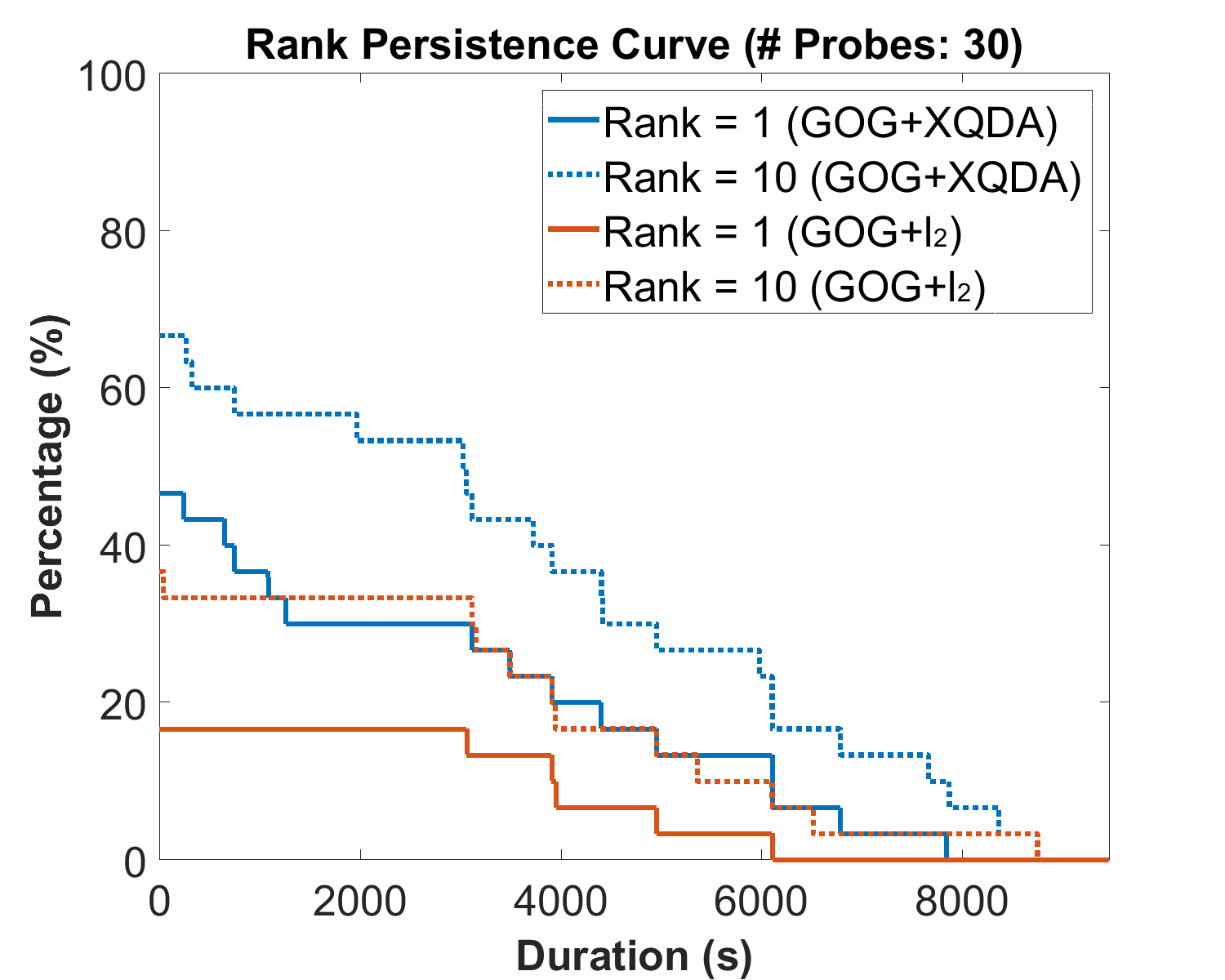}}%
		\qquad
		\subfloat[Probe camera 5, Gallery camera 1]{\includegraphics[width=0.35\linewidth]{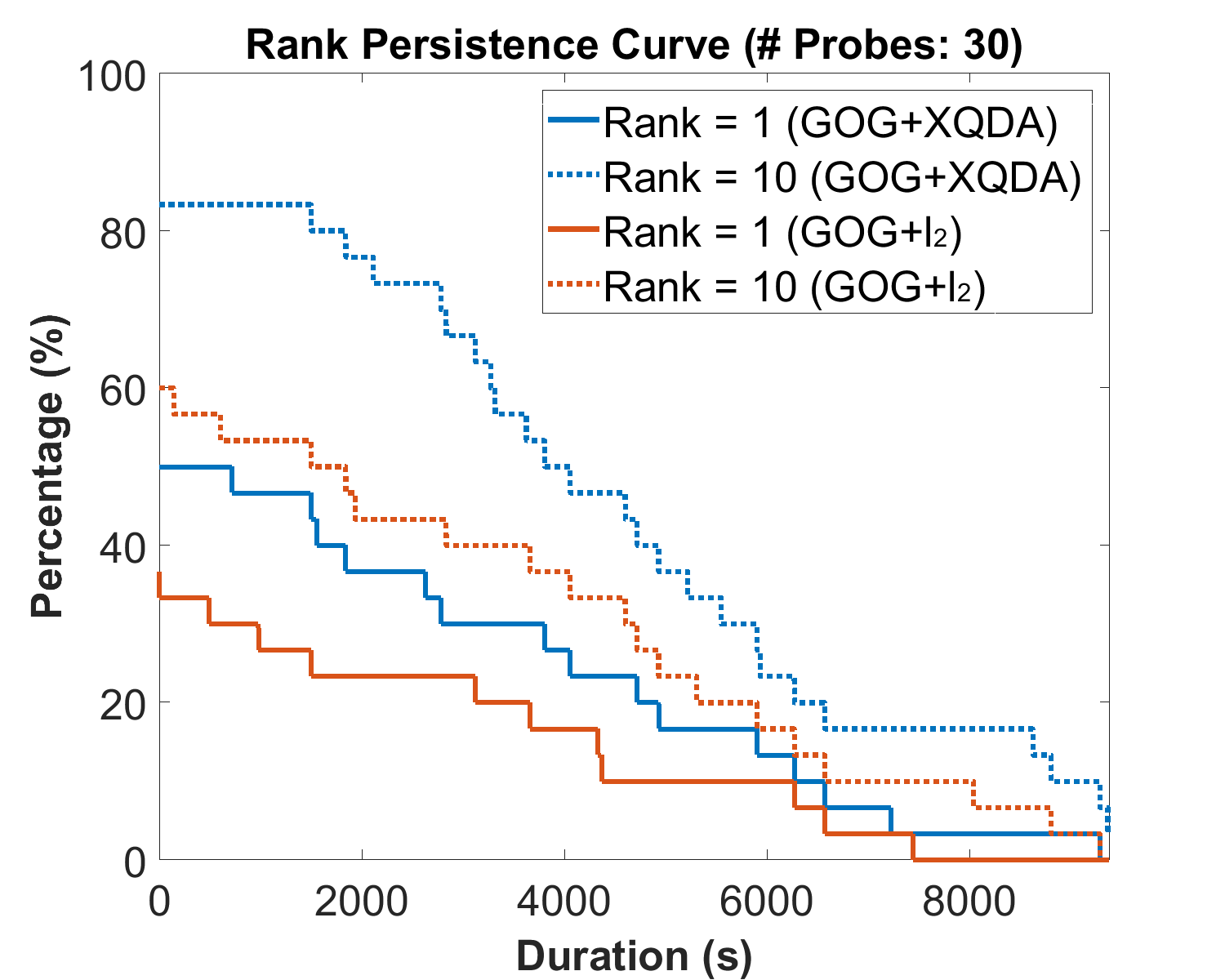}}%
\\
\subfloat[Probe camera 1, Gallery camera 3]{\includegraphics[width=0.35\linewidth]{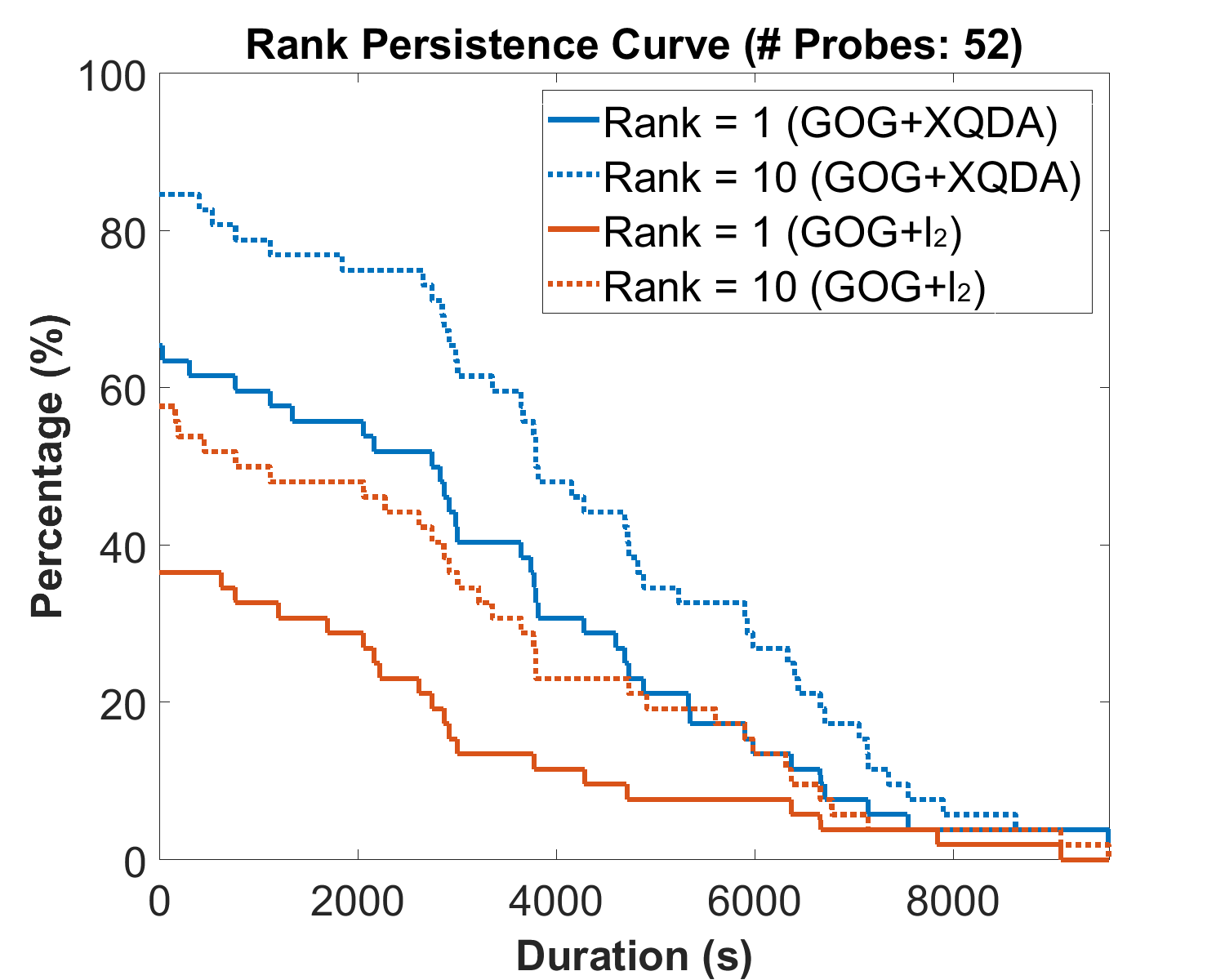}}%
		\qquad
		\subfloat[Probe camera 3, Gallery camera 1]{\includegraphics[width=0.35\linewidth]{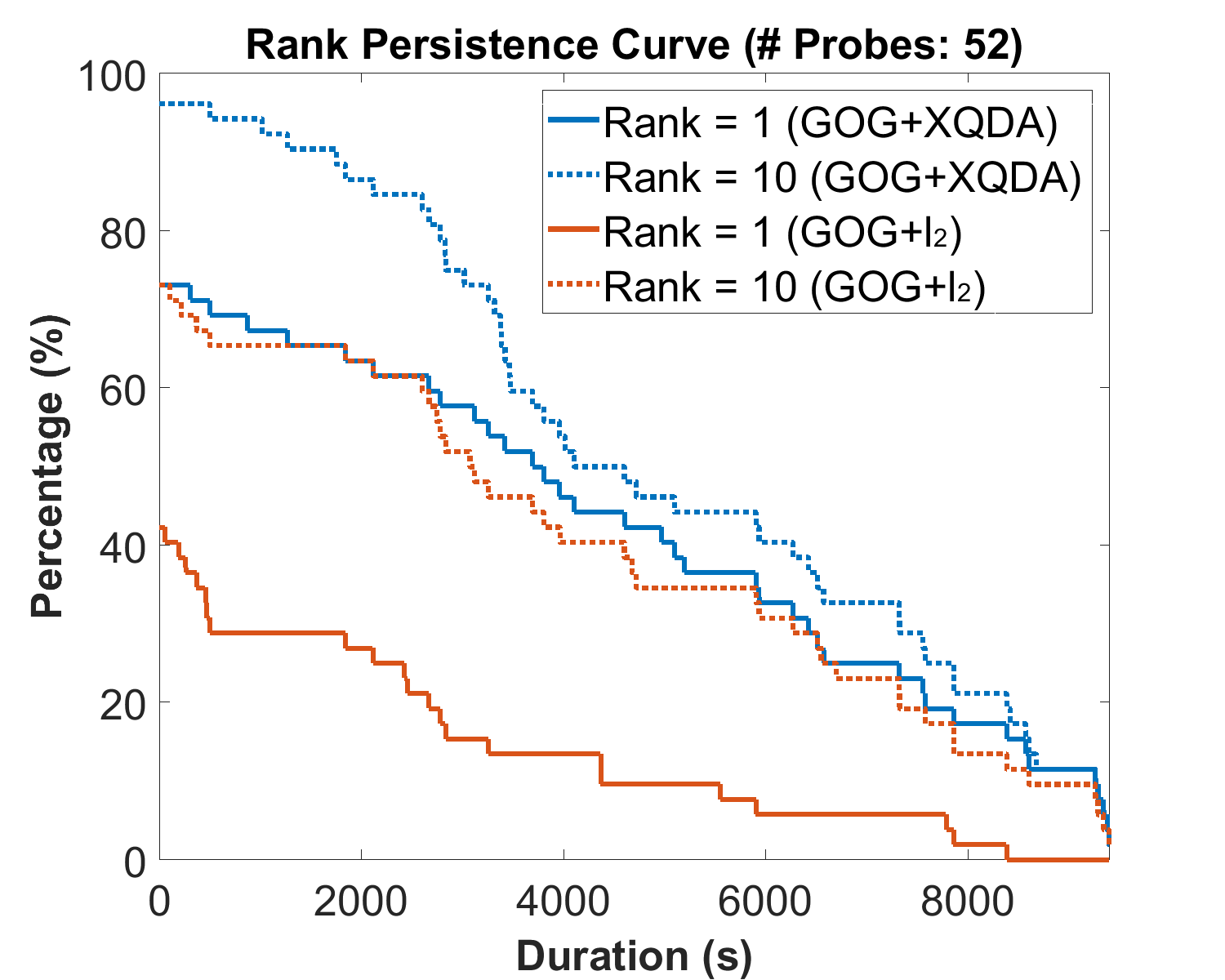}}%
		\caption{RPCs of multiple probes for camera pairs (1, 5) and (1, 3). GOG+XQDA is compared to the baseline with fixed ranks $r=1$ and $10$. `\# Probes' indicates the total number of probes considered in the RPC plot.}
		\label{fig:pair_cp3}
	\end{figure*}

	When evaluating the temporal performance of algorithms across multiple probes for cross-view re-identification, the same algorithm may produce different performance if the probe and gallery camera are interchanged. For different camera pairs, performance will also vary due to cross-view differences. In Figure \ref{fig:pair_cp3}, we present RPCs for 2 camera pairs. In plotting RPCs in Figure \ref{fig:pair_cp3}, we consider all participants who ever appeared in both cameras of the camera pair. For Figure \ref{fig:pair_cp3}(a), the probes are taken from Camera 1, whereas the candidates are taken from Camera 5, and vice versa for Figure \ref{fig:pair_cp3}(b). Figure \ref{fig:pair_cp3}(c) and (d) are plotted in a similar way for Cameras 1 and 3.

	First, let's focus on the RPCs for the same camera pair. Consider the camera pair (1, 5) for illustration purposes. While having exactly the same number of participants in Figure \ref{fig:pair_cp3}(a) and (b), the same re-id algorithm produces different performance in two different time-varying galleries. From Table \ref{table:stat}, we can see a larger number of distractors in the gallery set of Camera 5, which will likely degrade the performance shown in Figure \ref{fig:pair_cp3}(a) compared to (b). Furthermore, for the camera pair (1, 3), the graphs with the same color and linestyle in Figure \ref{fig:pair_cp3}(d) are also higher than (c), indicating higher performance when candidates are from the gallery of Camera 1. This indicates that a person of interest will more likely persist in a rank-1 or rank-10 shortlist for a longer time in Camera 1 when compared to Camera 3 or 5. We see that the RPC can thus also be used to derive conclusions regarding which camera views in a camera network are more likely to ``hold on'' to the person of interest, an aspect of crucial importance in sensitive surveillance applications. 
	
	More significantly, we can see an obvious performance difference in Figures \ref{fig:pair_cp3}(b) and (d) when the same gallery set is considered. One cause for the difference is that different sets of probe images are considered in Figure \ref{fig:pair_cp3}(b) and (d). Another reason is the viewpoint variations across the camera views of these camera pairs. As we discussed at the beginning of Section \ref{sec:Exp}, all positive pairs from different camera views are trained for XQDA learning simultaneously. Due to different illumination conditions or viewpoint changes, some camera pairs could be easier for the trained metric, reflected in better cross-view re-id performance.  Due to the unique location of Camera 5, there are substantially more shadows and illumination variation patterns compared to Cameras 1 and 3, resulting in the performance degradation shown in Figure \ref{fig:pair_cp3}(b). All these factors result in worse temporal performance of the re-id algorithm, which is reflected in the lower value of the RPCs for a fixed duration, shown in Figure \ref{fig:pair_cp3}(b). 
	
	\subsubsection{Comparison of Multiple Re-id Algorithms}
	With the proposed RPCs, we can now easily compare the temporal performance of competing re-id algorithms. In this section, we will compare the RPCs of different re-id algorithms for cross-view re-id on a specific camera pair. Specifically, we choose camera pair (10, 12) in our dataset. Additional results on other pairs are provided as part of the supplementary material. There are 50 participants who appear in both camera views, of which we randomly pick 25 for metric learning. 

For feature extraction, we consider GOG \cite{GOG_CVPR16}, LOMO \cite{LOMO_XQDA_CVPR15}, WHOS \cite{WHOS_PAMI15}, HistLBP \cite{HistLBP_kernel_ECCV14}, IDE-VGGNet \cite{VGGnet_CoRR14}, IDE-ResNet \cite{Resnet_CVPR16}, and IDE-CaffeNet \cite{Alexnet_NIPS12}. In IDECaffeNet, IDE-ResNet, and IDE-VGGNet, we use the idea presented in \cite{zheng2016person} by Zheng et al., in which every person is treated as a separate class and a convolutional neural network is trained with a classification objective. AlexNet \cite{Alexnet_NIPS12}, ResNet \cite{Resnet_CVPR16}, and VGGNet \cite{VGGnet_CoRR14} architectures are employed in IDE-CaffeNet, IDE-ResNet and IDE-VGGNet respectively. In each case, we start with a model pre-trained on the ImageNet dataset, and fine-tune it using training data from 14 existing benchmark datasets, as described more fully in  Karanam \etal \cite{benchmark_17}. 

\begin{figure*}[h!]
		\centering
        \begin{tabular}{ccc}
\includegraphics[width=0.31\linewidth]{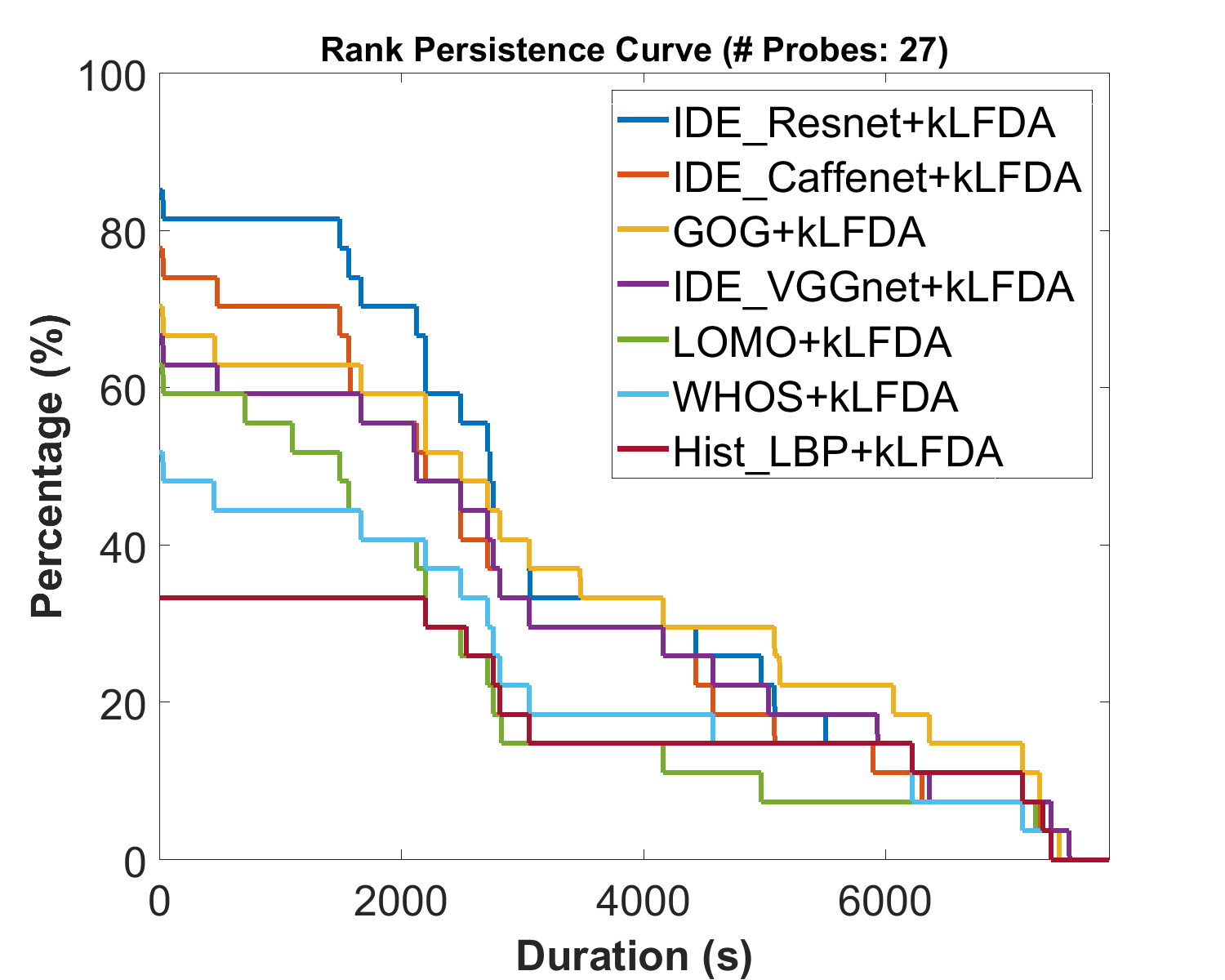} &
\includegraphics[width=0.31\linewidth]{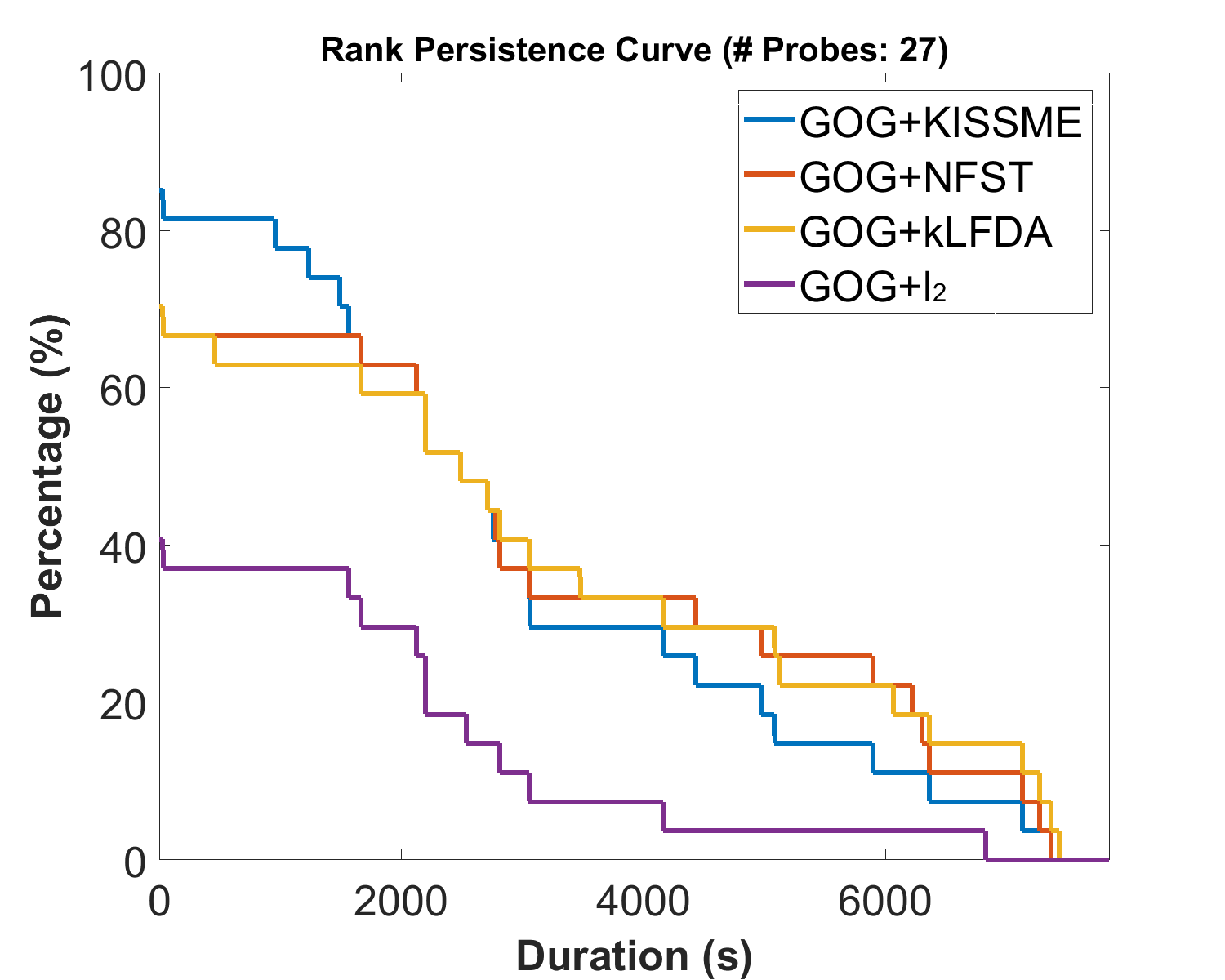} &
\includegraphics[width=0.31\linewidth]{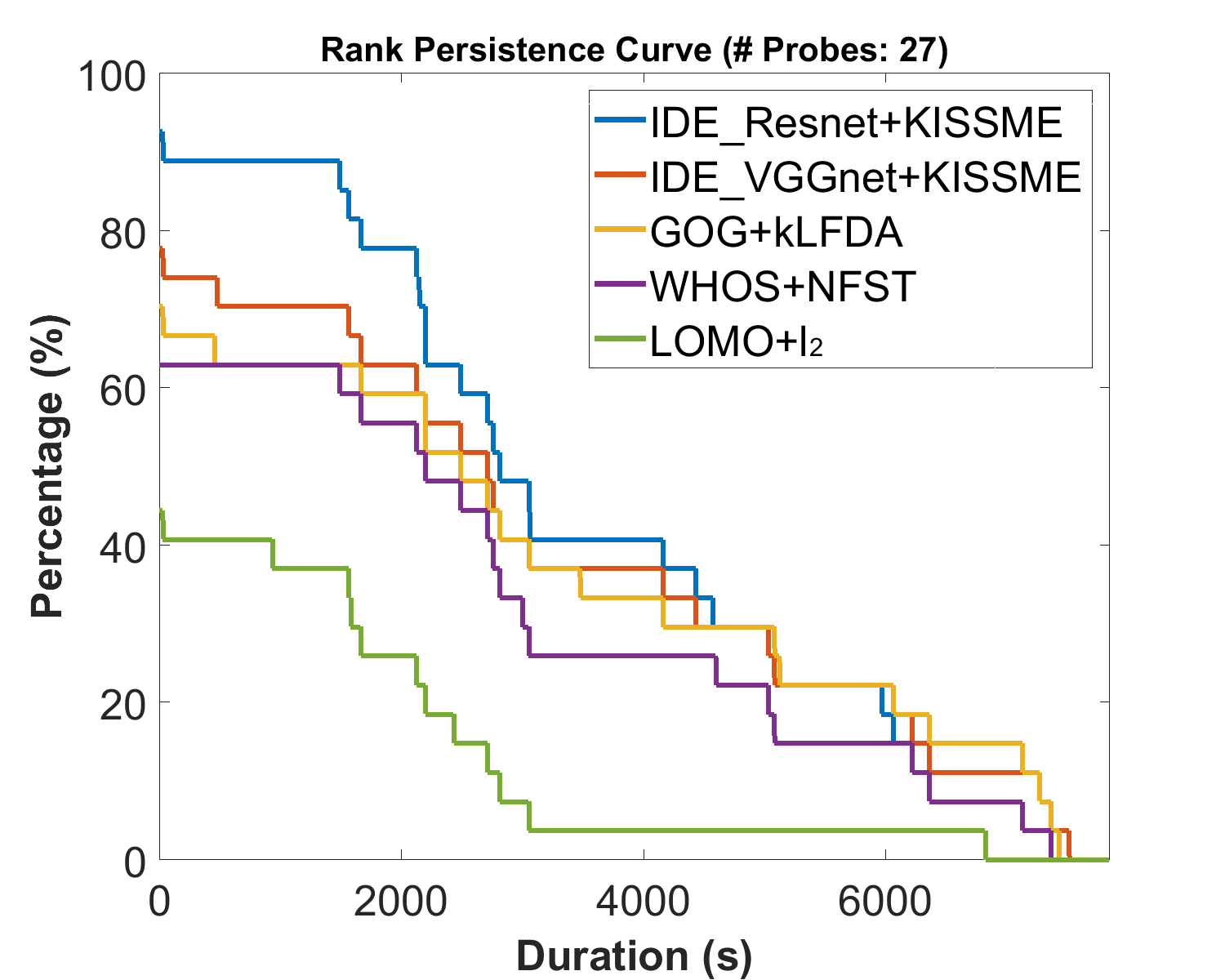} \\
(a) & (b) & (c) \\
\includegraphics[width=0.31\linewidth]{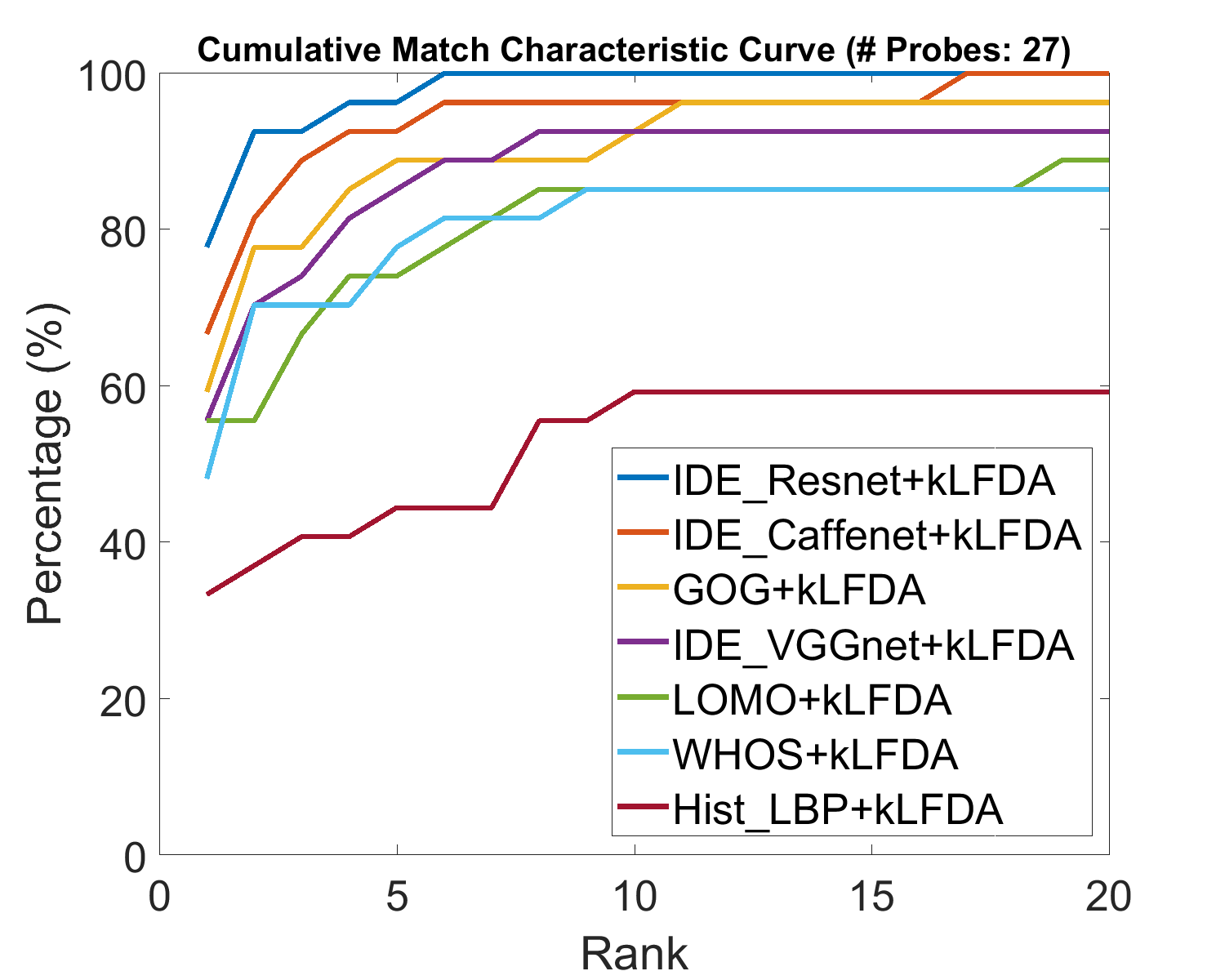} &
\includegraphics[width=0.31\linewidth]{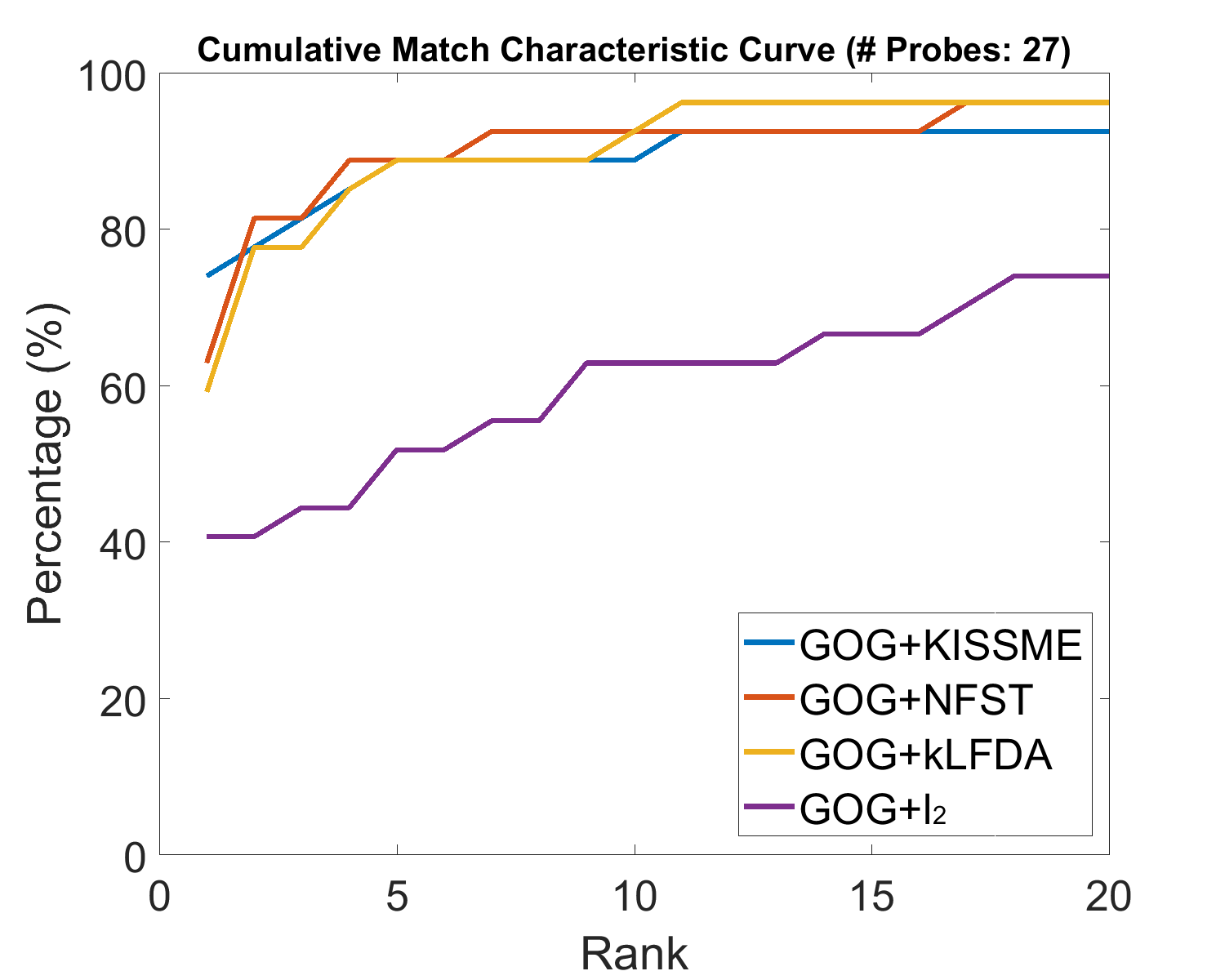} &
\includegraphics[width=0.31\linewidth]{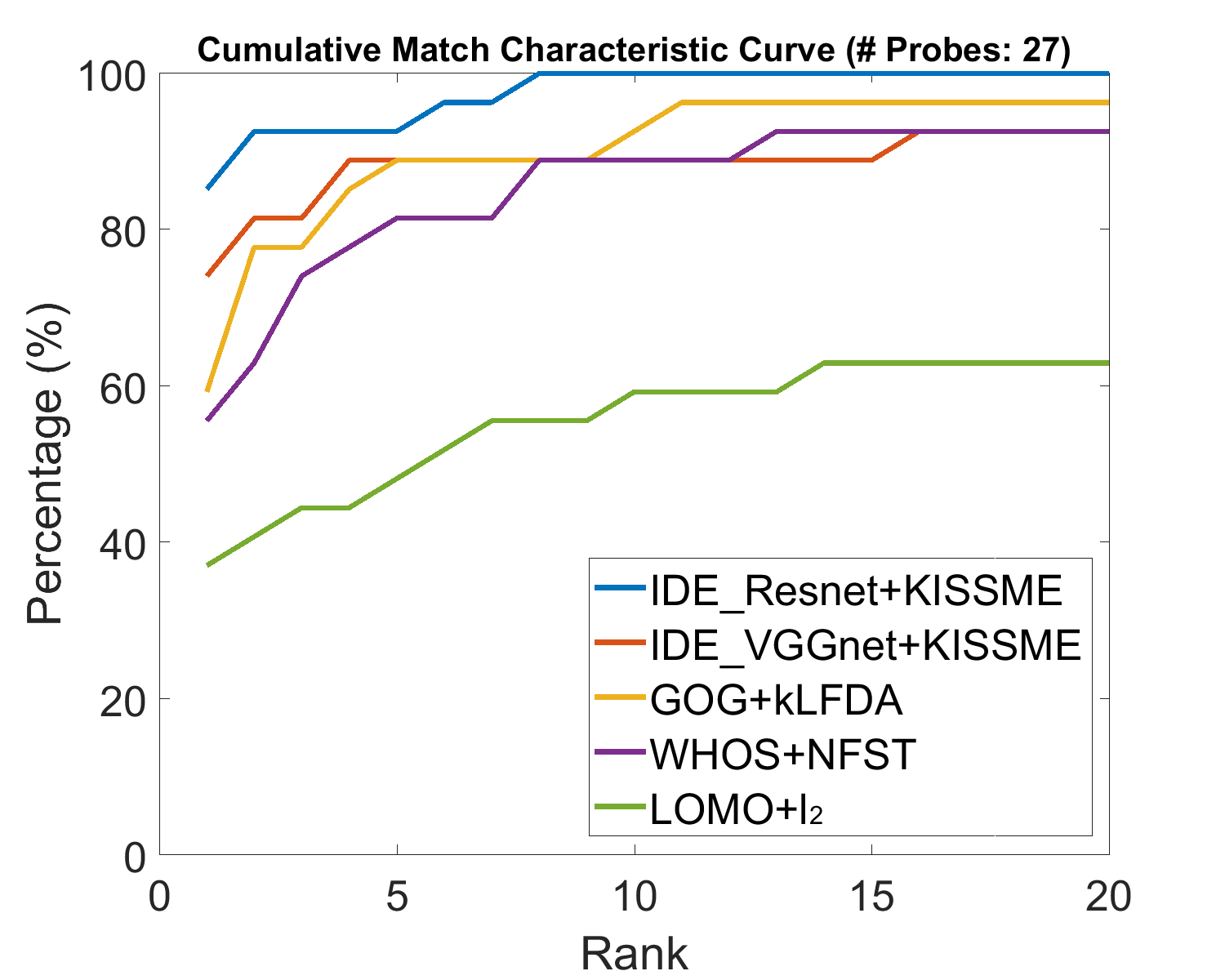} \\
(d) & (e) & (f) 
\end{tabular}
		\caption{Comparing rank 1 RPCs and CMCs of different re-id algorithm combinations applied to camera pair (10, 12). Camera 10 is the probe camera and Camera 12 is the gallery camera. (a) Rank-1 RPCs for kLFDA metric learning with different feature extraction methods. (b) Rank-1 RPCs for GOG feature extraction with different metric learning methods.  (c) Rank-1 RPCs for well-performing feature/metric combinations.  (d)-(f) CMC curves corresponding to the experiments in (a)-(c).}
		\label{fig:cp_algos}
	\end{figure*}

For metric learning, we consider KISSME \cite{KISSME_CVPR12}, XQDA \cite{LOMO_XQDA_CVPR15}, NFST (linear kernel) \cite{NFST_CVPR16} and kLFDA (linear kernel) \cite{LFDA_CVPR13,HistLBP_kernel_ECCV14}, algorithms that were shown to perform well in the study of Karanam \etal \cite{benchmark_17}. The RPCs for rank $r=1$ of various combinations of these re-id algorithms are shown in Figure \ref{fig:cp_algos} (a-c). The corresponding CMC curves are plotted in Figure \ref{fig:cp_algos} (d-f).

	In Figure \ref{fig:cp_algos}(a), for a fixed metric learning method, kLFDA, we can easily compare the temporal performance of various feature extraction methods. We note that IDE-ResNet achieves the best performance with IDE-CaffeNet, GOG and IDE-VGGNet not too far behind, with all these methods performing better than HistLBP. In Figure \ref{fig:cp_algos}(b), for a given feature extraction method, GOG, we compare the temporal performance of competing metric learning methods. While KISSME produce better results for short durations, kLFDA and NFST (linear kernel) generate better results at longer durations, with all of them performing much better than the baseline Euclidean distance ($l_2$) method. Figure \ref{fig:cp_algos}(c) gives a general comparison of re-id algorithms with different combinations of feature extraction and metric learning methods. For illustration purposes, we only select the top few. We can see that IDE-Resnet+KISSME, IDE-VGGnet+KISSME, and GOG+kLFDA achieve comparable results at mid-to-long durations, with IDE-Resnet+KISSME performing better for shorter durations. As we illustrated in previous sections, CMC curves in Figure \ref{fig:cp_algos}(d-f) can directly compare multiple algorithms as well, but are missing any operationally useful temporal comparison. 
	
	\section{Discussion and Future Work}

	Now that the underlying computer vision and machine learning technologies for re-id have matured, we propose that re-id researchers should begin to take a broader view of evaluating how re-id algorithms should integrate into functional real-world systems. For example, as noted in Li et al. \cite{li2014real} and Camps et al. \cite{Octavia_CSVT16}, the problem of comparing one candidate rectangle of pixels to another is only a small part of a fully automated re-id system. Instead, we must take into account that the candidate rectangles are likely generated by an automatic (and possibly inaccurate) human detection and tracking subsystem, and that the overall system needs to operate in real time. Instead of benchmarking datasets in which the gallery images are acquired only a few moments after the probe images, we should consider  applications such as crime prevention in which a perpetrator may return to the scene of the crime days after their initial detection. In such cases, the gallery of candidates is ever-expanding, and for certain periods of time may not contain the person of interest at all. 
	
	In this paper, we take a step towards this objective. To simulate a fully automated re-id system with human detection and tracking functionality, we collected a new multi-camera dataset with all time stamps preserved and all person images generated by the ACF human detector. The dataset contains multiple planned reappearances of participants and many distractors in order to simulate a real-world re-identification application such as crime prevention. Temporal analysis is then applied to this dataset to help us evaluate the performance of multiple re-id algorithms functioning in a simulated re-id system. While our experiments show interesting results, they also pose several challenges we wish to tackle in the future:
	\begin{itemize}
		\item As we stated, in real-world re-id applications such as crime prevention, a perpetrator may return to the crime scene after a long period of time, e.g., a day or more, during which the gallery would be filled with a huge amount of distractors. This poses a significant challenge to re-id algorithms in terms of how to correctly match the same person to a huge list of candidates. In our current dataset, however, the time spread of participants' reappearances is relatively short, typically within 2 hours, which is an approximate simulation of such real situations.  Extremely long duration data collections with known time-stamped reappearances would be very useful.
        
        \item A criminal might naturally change his or her appearance when returning to the same scene, for example wearing a hat or mask, which will also aggravate the difficulty of re-identification. In our dataset, some of the participants took off/carried their backpacks when entering different camera views, but more types of appearance change would be preferred in  future studies.
        
		\item Based on our proposed Rank Persistence Curve, the performance of competing re-id algorithms for a single camera pair can be visualized through an easy-to-read graph. More generally, we need to consider how to evaluate re-id algorithm performance for an entire multi-camera network. This raises interesting but challenging questions such as how to consider multiple galleries from different camera views at the same time, and how to construct a consistent evaluation scheme to integrate all these different galleries.
		\item Temporal considerations, in terms of both the ever-increasing gallery size and multiple probe appearances, lead to natural challenges from a feature and metric learning point of view. These can potentially be addressed by using temporally incremental approaches to learning re-id models \cite{temporal_ECCV16,HILP_CoRR16} where the model can be temporally adapted over time using either automated or human-in-the-loop feedback.
		\item In our experience with integrating academic re-id algorithms into operational surveillance command centers, we found the issue of user interfaces to be extremely important. The similarity between a rank-k shortlist and a police lineup was an effective analogy. However, the potentially very long time scales for crime prevention applications requires the re-evaluation of an operationally meaningful shortlist. Should candidates ``age out" of the ranked list using some sort of forgetting factor or re-ranking scheme \cite{Rerank_CVPR15,Rerank_ICCV15}? Should extremely promising candidates from long ago be kept alongside less-certain but more timely recent candidates? Should the time-varying gallery contain all candidates ever seen or only those from the last N minutes? These considerations require close consultation with the potential users of the system to understand and set expectations and corresponding interface choices.
	\end{itemize}
	
	{\small
		\bibliographystyle{IEEEtran}
		\bibliography{ref}
	}


\begin{IEEEbiography}[{\includegraphics[width=1in,height=1.25in,clip,keepaspectratio]{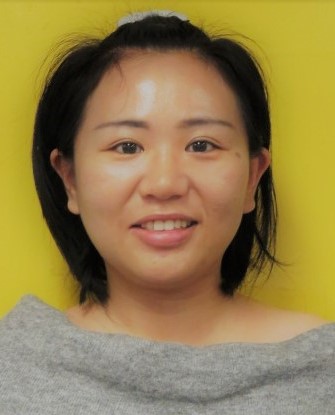}}]{Meng Zheng}
Meng Zheng is a Ph.D.~student in Electrical, Computer \& Systems Engineering from Rensselaer Polytechnic Institute. She received a M.S. and B.Eng. degree in School of Information and Electronics from Beijing Institute of Technology, China. Her research interests include computer vision and machine learning with a focus on person re-identification.
\end{IEEEbiography}

\begin{IEEEbiography}
[{\includegraphics[width=1in,height=1.25in,clip,keepaspectratio]{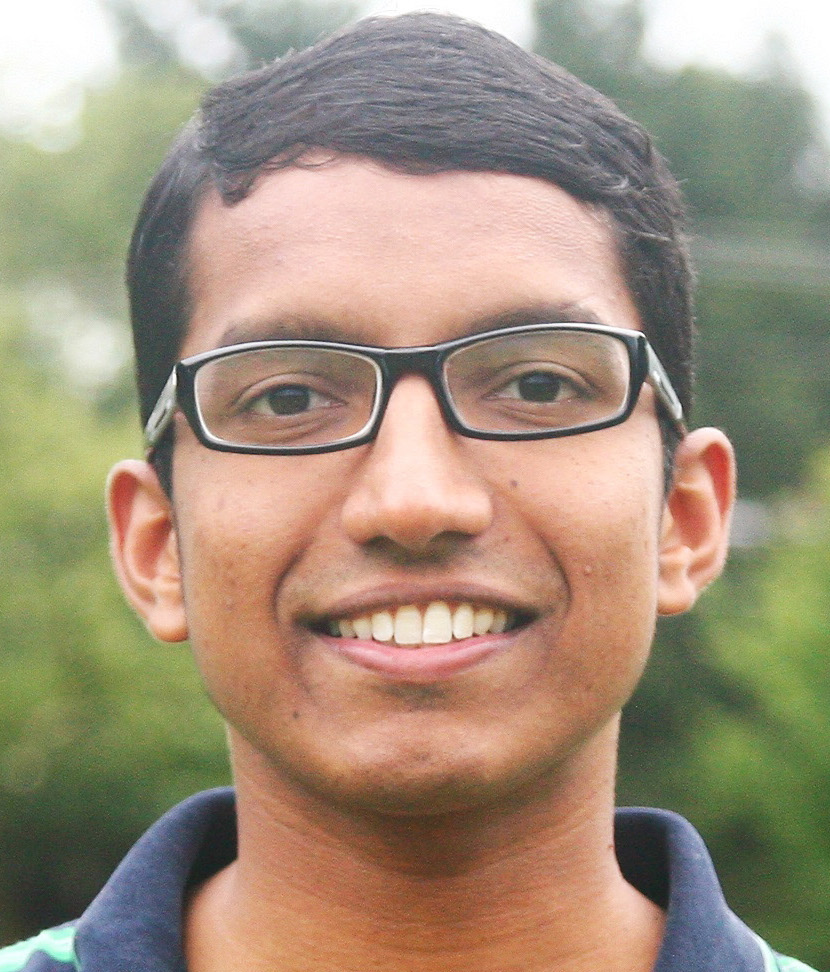}}]{Srikrishna Karanam}
Srikrishna Karanam is a Research Scientist in the Vision Technologies and Solutions group at Siemens Corporate Technology, Princeton, NJ. He has a Ph.D.~degree in Computer \& Systems Engineering from Rensselaer Polytechnic Institute. His research interests include computer vision and machine learning with a focus on all aspects of image indexing, search, and retrieval for object recognition applications.
\end{IEEEbiography}

\begin{IEEEbiography}[{\includegraphics[width=1in,height=1.25in,clip,keepaspectratio]{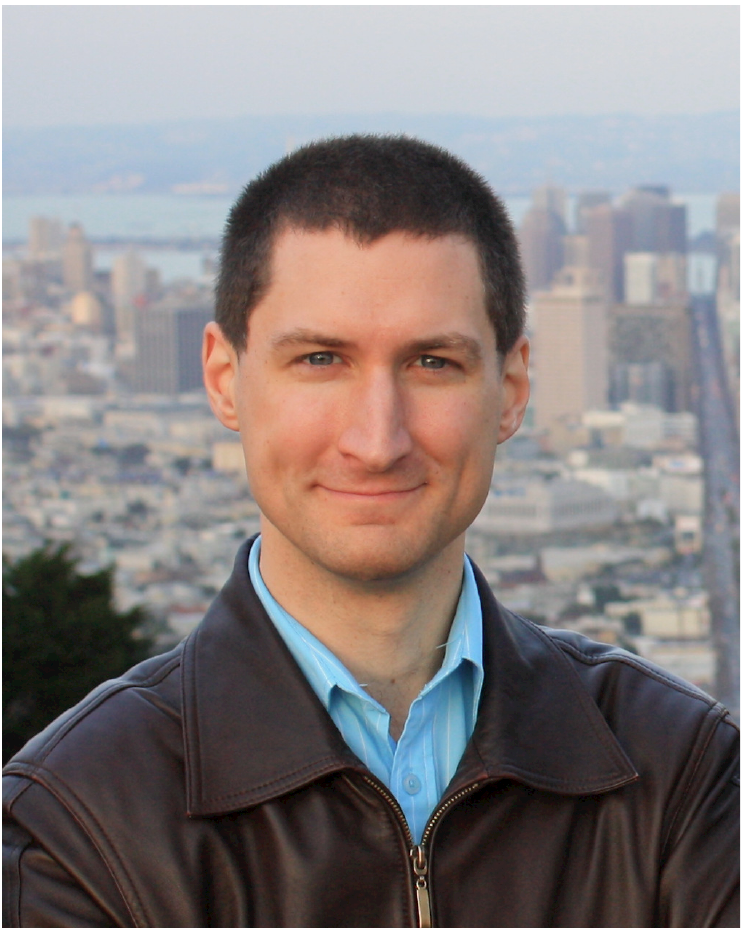}}]{Richard J. Radke}
Richard J.~Radke joined the Electrical, Computer, and Systems Engineering department at Rensselaer Polytechnic Institute in 2001, where he is now a Full Professor. He has B.A.~and M.A.~degrees in computational and applied mathematics from Rice University, and M.A.~and Ph.D.~degrees in electrical engineering from Princeton University.  His current research interests involve computer vision problems related to human-scale, occupant-aware environments, such as person tracking and re-identification with cameras and range sensors.  Dr.~Radke is affiliated with the NSF Engineering Research Center for Lighting Enabled Systems and Applications (LESA), the DHS Center of Excellence on Explosives Detection, Mitigation and Response (ALERT), and Rensselaer's Experimental Media and Performing Arts Center (EMPAC). He received an NSF CAREER award in March 2003 and was a member of the 2007 DARPA Computer Science Study Group.  Dr.~Radke is a Senior Member of the IEEE and a Senior Area Editor of \emph{IEEE Transactions on Image Processing}.  His textbook \emph{Computer Vision for Visual Effects} was published by Cambridge University Press in 2012.
\end{IEEEbiography}
	
\end{document}